%% file: _main.tex
\pdfoutput=1

\documentclass{article}

\usepackage[preprint]{neurips_2023}

\usepackage[utf8]{inputenc} %
\usepackage[T1]{fontenc}    %
\usepackage{hyperref}       %
\usepackage{url}            %
\usepackage{booktabs}       %
\usepackage{amsfonts}       %
\usepackage{nicefrac}       %
\usepackage{microtype}      %
\usepackage{xcolor}         %

\usepackage{svg}

\input{00_macro.tex}

\title{Understanding Variation in Subpopulation Susceptibility to Poisoning Attacks}

\author{%
Evan Rose, Fnu Suya, David Evans\\
Department of Computer Science\\
University of Virginia\\
\texttt{\{esr6vqa,suya,evans\}@virginia.edu}\\
}

\begin{document}

\maketitle

\input{00_abstract}

\input{01_introduction}

\input{02_preliminary}

\input{03_synthetic_experiments}

\input{04_adult_experiments}

\input{05_limitation}

\input{06_conclusion}

\setcitestyle{numbers,square,comma}
\bibliographystyle{plain}
\bibliography{references}

\end{document}

%% file: 00_macro.tex
\usepackage{amssymb}
\usepackage{amsmath}
\usepackage{bm}
\usepackage{bbm}
\usepackage{amsthm}
\usepackage{mathtools}
\usepackage[capitalize,noabbrev]{cleveref}
\usepackage{wrapfig}
\usepackage{siunitx}
\usepackage{enumitem}
\setenumerate{noitemsep,topsep=0pt,parsep=0pt,partopsep=0pt}
\usepackage[utf8]{inputenc} %
\usepackage[T1]{fontenc}    %
\usepackage{url}            %
\usepackage{booktabs}       %
\usepackage{amsfonts}       %
\usepackage{nicefrac}       %
\usepackage{microtype}      %
\usepackage{todonotes}
\usepackage{multirow}
\usepackage{xcolor}         %

\usepackage{graphicx}

\usepackage{subcaption}
\usepackage{dsfont}

\usepackage{siunitx}

\theoremstyle{plain}

\theoremstyle{remark}

\theoremstyle{definition}

\newcommand\shortsection[1]{\vspace{6pt}{\noindent\bf #1.}}

\usepackage{color}
\usepackage{todonotes}
\definecolor{ForestGreen}{cmyk}{0.864, 0.0, 0.429, 0.396}
\newcommand{\ifcomments}{\iftrue}

\DeclareMathOperator*{\argmin}{argmin}

\newcommand{\cS}{\mathcal{S}}

\newcommand{\cX}{\mathcal{X}}
\newcommand{\cY}{\mathcal{Y}}

\newcommand{\cP}{\mathcal{P}}

\newcommand{\RR}{\mathbb R}

\newcommand{\ind}{\mathbbm{1}}

\newcommand{\bx}{\bm{x}}
\newcommand{\bw}{\bm{w}}

\newcommand{\sgn}{\mathrm{sgn}}

%% file: 00_abstract.tex
\begin{abstract}
 Machine learning is susceptible to poisoning attacks, in which an attacker controls a small fraction of the training data and chooses that data with the goal of inducing some behavior unintended by the model developer in the trained model. We consider a realistic setting in which the adversary with the ability to insert a limited number of data points attempts to control the model’s behavior on a specific subpopulation. Inspired by previous observations on disparate effectiveness of random label-flipping attacks on different subpopulations, we investigate the properties that can impact the effectiveness of state-of-the-art poisoning attacks against different subpopulations. For a family of 2-dimensional synthetic datasets, we empirically find that dataset separability plays a dominant role in subpopulation vulnerability for less separable datasets. However, well-separated datasets exhibit more dependence on individual subpopulation properties. We further discover that a crucial subpopulation property is captured by the difference in loss on the clean dataset between the clean model and a target model that misclassifies the subpopulation, and a subpopulation is much easier to attack if the loss difference is small. This property also generalizes to high-dimensional benchmark datasets. For the Adult benchmark dataset, we show that we can find semantically-meaningful subpopulation properties that are related to the susceptibilities of a selected group of subpopulations. The results in this paper are accompanied by a fully interactive web-based visualization of subpopulation poisoning attacks found at \url{https://uvasrg.github.io/visualizing-poisoning/}.

\end{abstract}

%% file: 01_introduction.tex
\section{Introduction}\label{sec:intro}
Machine learning models are often trained on data points collected from untrusted third parties~\citep{carlini2023poisoning}. Training models on these potentially malicious data points poses security risks. A typical application is in spam filtering, where the spam detector is trained using data (i.e., emails) that are generated by users with labels often provided implicitly by user actions. In this setting, spammers can generate spam messages that inject benign words likely to occur in spam emails such that models trained on these spam messages will incur significant drops in filtering accuracy as benign and malicious messages become indistinguishable~\citep{nelson2008exploiting,huang2011adversarial}. These kinds of attacks are known as \emph{poisoning attacks}. In a poisoning attack, the attacker injects a relatively small number of crafted examples into the training set such that the resulting trained model (known as the \emph{poisoned model}) performs in a way that satisfies the attacker's goals. 

Most research on poisoning attacks focuses on one of two extreme poisoning goals: \emph{indiscriminate poisoning attacks}, where the adversary’s goal is simply to decrease the overall accuracy of the model \citep{biggio2011support,biggio2012poisoning,xiao2012adversarial,mei2015security,mei2015using,steinhardt2017certified,koh2022stronger} and \emph{targeted poisoning attacks}, where the goal is to induce a classifier that misclassifies a particular known input~\citep{shafahi2018poison,zhu2019transferable,koh2017understanding,geiping2020witches,huang2020metapoison}. Jagielski et al.~\citep{jagielski2019subpop,jagielski2021subpopulation} introduced a more realistic attacker objective known as a \emph{subpopulation poisoning attack}, where the goal is to increase the error rate or obtain a particular output for a defined subset of the data distribution. Their experiments predominantly rely on (relatively weak) random label-flipping attacks evaluated on a few selected subpopulations. In these experiments, they observe the existence of disparate vulnerabilities, measured by the absolute increase in test errors on the subpopulation, across selected subpopulations, with some subpopulations being more vulnerable to poisoning than others. This raises an interesting question on whether the properties of the individual subpopulations have an impact on the effectiveness of state-of-the-art poisoning attacks, which has not been explored in previous works. Additionally, one might question whether the empirical observation using the weaker random label-flipping attack holds for state-of-the-art poisoning attacks that can manipulate both the data features and labels. 

The goal of this work is to systematically analyze the possible factors of the subpopulations that impact the attack effectiveness of state-of-the-art subpopulation poisoning attacks. A preliminary version of the results in this paper was presented as an interactive data visualization at the VISxAI workshop that focuses on explaining AI through visualizations~\citep{evan2022poisoning}.
We encourage readers to visit \url{https://uvasrg.github.io/visualizing-poisoning/} to explore the visualizations described here. This paper extends the analysis through visualization by testing with additional state-of-the-art subpopulation attacks \citep{koh2022stronger} (require target models as inputs) and evaluating them with a greater number of carefully selected target models. 

\shortsection{Contributions}
To further confirm Jagielski et al.'s observation~\citep{jagielski2019subpop,jagielski2021subpopulation} regarding the disparate vulnerability of subpopulations, we empirically measure the subpopulation susceptibilities on a larger number of subpopulations for both synthetic datasets (\Cref{sec:subpoplulation synthetic experiment}) and the Adult benchmark (\Cref{sec:subpoplulation adult experiment}) dataset. In particular, in \Cref{sec:synthetic subpopulation variability}, we confirm the disparate vulnerability of subpopulations within synthetic datasets by conducting tests on a large number of subpopulations using state-of-the-art poisoning attacks. We show that for datasets that are not separable, subpopulations are generally all vulnerable to poisoning with only minor variation in vulnerability across subpopulations. 
Conversely, in well-separated datasets, subpopulation-specific properties such as relative position  take precedence in determining susceptibility. In \Cref{sec:synthetic subpopulation property}, we further demonstrate that the attack effectiveness of the state-of-the-art attacks is highly correlated with the subpopulation property that is measured by the model loss difference between the clean model and a target model that misclassifies the subpopulation. Then, in \Cref{sec:subpoplulation adult experimet results}, we demonstrate that a significant spread in subpopulation vulnerability still exists for the benchmark Adult dataset when evaluated against state-of-the-art poisoning attacks across a large number of subpopulations. We find that the generally descriptive factor of model loss difference still correlates highly with the attack effectiveness on the benchmark dataset. Finally, we also show that there can be some semantically meaningful properties that are related to attack effectiveness when only selected subpopulations are measured, but generalizing the semantically meaningful properties is hard (\Cref{sec:subpoplulation adult experimet semantic}).  

\shortsection{Related Work}
The closest related work is the work on subpopulation attacks by Jagielski et al.~\citep{jagielski2019subpop,jagielski2021subpopulation}, who observed disparate poisoning effectiveness against subpopulations from the UCI Adult \citep{Dua:2019} and UTKFace \citep{zhang2017age} datasets under their proposed attack framework  and discovered that accuracy reduction difference can as huge 
as 44\% for the tested 10 UCI Adult subpopulations. We confirm their initial observation of disparate vulnerabilities among subpopulations by testing on a larger number of subpopulations using state-of-the-art poisoning attacks, and also identifying distributional and subpopulation properties that are highly related to the empirical vulnerabilities. 

Poisoning attacks have also been directly applied to algorithmic fairness \citep{chang2020adversarial, solans2020poisoning}. These works primarily focus on poisoning \textit{fair} classifiers, which are required to satisfy certain technical criteria such as equalized odds \citep{hardt2016equality} or disparate impact \citep{barocas_big_2016}. While our work can be considered in the context of algorithmic fairness, we do not study classifiers under fairness restrictions. Our goal is to understand the vulnerability of subpopulations to poisoning attacks in a more general context for which fairness may not be a primary concern (e.g., malware detection or spam filtering).

Different subpopulation poisoning attack algorithms are proposed \citep{jagielski2019subpop,jagielski2021subpopulation,suya2021model} or can be easily adapted from the indiscriminate poisoning settings \citep{koh2022stronger}. Our work leverages these attacks as the tools to empirically assess the vulnerability of the subpopulations, but focuses on identifying the possible factors of the subpopulations that are relevant to the disparate poisoning effectiveness. Another work that studies the inherent vulnerabilities of individual test samples under targeted poisoning attacks may also be applied for the subpopulation setting \citep{wang2022lethal}. However, a direct extension of targeted attacks to subpopulation settings might overestimate the power of subpopulation attacks and cannot well reflect the inherent vulnerabilities of subpopulations to poisoning.

%% file: 02_preliminary.tex
\section{Preliminaries}\label{sec:preliminary}
We consider binary classification tasks. 
Let $\cX\subseteq\RR^n$ be the input space and $\cY=\{-1, +1\}$ be the label space. We focus on the linear hypothesis class $\mathcal{H}$ and hinge loss, which is a common setting considered in prior works~\citep{biggio2011support,biggio2012poisoning,steinhardt2017certified,koh2022stronger,suya2021model}. A \emph{linear hypothesis} parameterized by a weight parameter $\bw\in\RR^n$ and a bias parameter $b\in\RR$ is defined as: $h_{\bw,b}(\bx) = \sgn(\bw^\top\bx + b) \text{ for any } \bx\in\RR^n$, where $\sgn(\cdot)$ denotes the sign function. 
For any $\bx\in\cX$ and $y\in\cY$, the \emph{hinge loss} of a linear classifier $h_{\bw, b}$ is defined as: 
\begin{align}
\label{eq:hinge loss}
\ell(h_{\bw,b}; \bx, y) = \max\{0, 1 - y(\bw^\top\bx+b)\} + \frac{\lambda}{2}\|\bw\|_2^2,
\end{align}
where $\lambda \geq 0$ is the tuning parameter which penalizes the $\ell_2$-norm of the weight parameter $\bw$. 

Given a clean training data $\mathcal{S}_c$ and a poisoned data $\cS_p$ (can be empty), the victim minimizes the following loss function to train a linear SVM model:
\begin{equation}
\label{eq:poisoned model}
    h_p = \argmin_{h \in \mathcal{H}} \frac{1}{\lvert \cS_c\cup \cS_p \rvert} L(h; \cS_c\cup \cS_p) + \frac{\lambda}{2}\|\bw\|_2^2,
\end{equation}
where $L(h; \cS)=\sum_{(\bm{x},y)\in \cS\cup \cS_p}\ell(h;\bm{x},y)$. 

\subsection{Subpopulation Data Poisoning Attacks}
\label{sec:subpopulation poisoning attacks}
We consider a slightly modified version of the subpopulation poisoning attack framework developed in Jagielski et al. \citep{jagielski2019subpop, jagielski2021subpopulation}. For simplicity, we describe subpopulations as subsets of the input space $\cX$. While the original definition is more general and defines subpopulations in terms of auxiliary information, it includes our definition as a special case. While this means that our experiments can technically be viewed as targeted attacks against a set of points, conceptually the subpopulation framework is more appropriate due to the techniques used to generate the subpopulations and the blindness to attacking specific points in the subpopulation, while targeted attacks generally aim to misclassify a particularly specified set of points.

For a clean dataset $\cS_c$, a target subpopulation $\cP \subseteq \cS_c$, a target label $y_t \in \cY$, and a success threshold $0 \le r \le 1$, we say that a set of poisoned datapoints $\cS_p$ achieves the subpopulation attack objective if
\begin{equation}
\label{eq:attack success}
    \frac{1}{|\cP|} \sum_{\bx \in \cP} \ind(h_p(\bx) = y_t) \ge r
\end{equation}
where $h_p$ is the classifier trained by the victim as defined as in \Cref{eq:poisoned model}.

\subsection{Poisoning Strategy and Evaluation Metrics}
\label{sec:subpopulation poison strategy and metric}
In this section, we first define subpopulation susceptibility and show how to measure the subpopulation susceptibility empirically given a poisoning attack. Then, we provide details on our choice of poisoning attack algorithms: the model-targeted poisoning (MTP) attack~\citep{suya2021model} and the KKT attack~\citep{koh2022stronger}, and explain how we use the attacks to obtain an empirical estimation of the subpopulation susceptibility. 

\shortsection{Empirically Measuring Subpopulation Susceptibility}
Given a clean dataset $\cS_c$, a subpopulation $\cP \subseteq \cS_c$, a target label $y_t \in \cY$, and a success threshold $0 \le r \le 1$, we define the \textit{difficulty} of the subpopulation poisoning attack against $\cP$ to be the ratio $|\cS_p| / |\cS_c|$ of the size of the smallest poisoning set $\cS_p$ that achieves the attacker objective (as defined in \Cref{eq:attack success}) to the size of the clean dataset $\cS_c$. Given a successful empirical attack against a subpopulation, an empirical estimate of the difficulty is given by using the size of the poisoning set used (or the smallest such set if several successful attacks are considered). Note that under our definition, subpopulations that are incorrectly classified by the clean classifier are trivially ``vulnerable'' in the sense that a vacuous attack without using any poisoning points already achieves the attacker objective. %
To restrict our view to only subpopulations of practical interest, we exclude subpopulations that already exceed the error rate objective from our analysis, including only those that require at least one poisoning point to achieve the attack objective. %

Our definition of attack difficulty stems from a desire to offer well-defined comparisons between different subpopulations in terms of their susceptibility scores. In particular, measuring the subpopulation vulnerability using the number of poisoning points needed to induce some fixed error threshold on the subpopulation can be visually clearer when comparing different attacks using visualization techniques. There are other ways to define the vulnerability of a subpopulation to poisoning attacks, such as the maximum achievable test error on the subpopulation as a function of the poisoning fraction. In this case, attack success is measured by a (close to) continuous error measurement (rather than the 0--1 binary success measure in our formulation). Such alternatives are outside the scope of this work, and we leave exploration of other vulnerability definitions to future work.

\shortsection{Simulated Attacker Objective}
Our simulated attacker's objective is to induce a classifier that misclassifies at least $r=50\%$ of the target subpopulation. We choose the 50\% threshold to mitigate the impact of outliers in the subpopulations and because empirically the 50\% threshold reasonably captures the essential properties of attacks against the subpopulations. In earlier experiments requiring 100\% attack success, we observed that attack difficulty was often determined by a few outliers in the test subpopulation. Since our eventual goal is to characterize attack difficulty in terms of the properties of the targeted subpopulation (which outliers do not necessarily satisfy), targeting at 50\% threshold seems like a reasonable relaxation.

\shortsection{Selected Poisoning Attacks}
We choose the model-targeted poisoning (MTP) attack in Suya et al.\ \citep{suya2021model} and the KKT attack in Koh et al.\ \citep{koh2022stronger} as the poisoning strategy due to their state-of-the-art performance on subpopulation poisoning attacks. These two attacks require a target model encoding some certain attacker objective as the input and then generate the poisoning points to induce the target model and ultimately, the encoded attacker objective. Suya et al.\ \citep{suya2021model} showed that the MTP attack can outperform the random label-flipping attack used in Jagielski et al.'s experiments \citep{jagielski2019subpop,jagielski2021subpopulation}. In addition, the gradient attacks, which were adapted to subpopulation settings with efficient relaxation by Jagielski et al.\ \citep{jagielski2021subpopulation}, can also get stuck in poor local optima, causing suboptimal performance \citep{koh2022stronger}. Therefore, we choose the MTP attack and the KKT attack, which avoids the problem of suboptimal solutions when the provided target model is sufficiently good. Below, we show how to generate good candidate target models and also how to run the chosen poisoning attacks with the generated target models.

\shortsection{Generation of Target Models}
The selected poisoning attacks all require a proper target model as input. Note that our attack objective is not, strictly speaking, a model-targeted objective, yet we still use the model-targeted framework for its state-of-the-art performance on subpopulation poisoning attacks \cite{suya2021model}. Since we do not know the best target models beforehand, a suitable approach is to run the attack against a diverse set of target models and pick the best performing attack (i.e., attack that uses fewest poisoning points to achieve the attacker objective). Following the target model generation method as in Suya et al. \citep{suya2021model,suya2023linear}, we generate target models of different errors on the subpopulation ranging from 50\% to 100\%, with a 5\% increment. The choice of the target models have significant impact on the effectiveness of the respective attacks in achieving the desired attack goals (e.g., achieving 50\% error on the subpopulation in our case) and therefore, we ensure the generated target models have the lowest loss on the clean training data while still achieving certain error rates on the subpopulations. This generation strategy is theoretically justified in Suya et al.~\citep{suya2021model}. We visually demonstrate the dependence of attack performance on the choice of target model in \Cref{sec:subpoplulation synthetic experimet setup}. 

\shortsection{Running the MTP and KKT with Target Models}
We run the MTP and KKT attack with variation on the practice by Suya et al.\ \citep{suya2021model}. For each target model generated from random label-flipping, similar to Suya et al.\ \citep{suya2021model}, we first run the MTP attack with the generated targets and terminate the algorithm when the induced model $h_p$ misclassifies at least $r=50\%$ of the target subpopulation measured on the test set. Note that this stop condition is used regardless of the attack objective used to generate the target model (e.g., 100\% test error on the subpopulation). Then, different from prior practice, we additionally perform a second phase of the MTP attack using the induced model $h_p$ as the target model. The second phase serves the purpose of producing a potentially more efficient version of the attacks in the first phase and also to diversify the set of target models produced. 

For each MTP attack in both phases, a set $\cS_p$ of poisoning points with size $n$ and a lower bound $lb$\footnote{The lower bound means there are no poisoning attacks that can induce the given target model by using fewer poisoning points than the given lower bound.} on the size of the poisoning set required to induce the respective target model are recorded. Finally, in the third phase, we run the KKT attack on both the set of label-flipped target models and the set of induced target models from the first phase. Note that the KKT attack also requires inputting the size of the poisoning set in advance, which is unknown in practice. To provide feasible poisoning set size, we choose five equally-spaced values in the interval $[lb, n)$, where we did not run the KKT attack on size $n$ because we already have a successful MTP attack at that size. The three phases of attacks are designed to better estimate the attack difficulty on subpopulations empirically, at the cost of increased computational cost compared to the efficient (but weak) random label-flipping attacks used by Jagielski et al.~\citep{jagielski2019subpop,jagielski2021subpopulation}.

The effectiveness of each stage of the attack methodology is reported in \Cref{tab:attack performance}. In both dataset settings, the most efficient attack was most often produced by the first round of attacks (denoted MTP-1 in the table) against target models generated from label-flipping. In a significant number of attacks against synthetic datasets, the second phase of the MTP attack (denoted MTP-2 in the table) produced the smallest successful poisoning set. In attacks against the Adult dataset, the second MTP phase produced the most efficient poisoning set only three times. We do not yet know the reason behind the discrepancy between the Adult and benchmark datasets, and future investigation may further clarify desirable properties of target models for state-of-the-art poisoning attacks. The KKT attack never produced the most efficient set of poisoning points, providing empirical evidence that (MTP) attacks in the first two phases are better estimations of the (inherent) attack difficulties compared to the KKT attacks in the third phase.

\begin{table*}[tb]
\centering
\caption[Best Attack Frequency]{The frequency of optimal attacks resulting from each phase of the attack methodology.}
\begin{tabular}{llll}
\hline
\textbf{Dataset} & \textbf{MTP-1} & \textbf{MTP-2} & \textbf{KKT} \\ \hline
Synthetic        & 8,581          & 3,736          & 0            \\ \hline
Adult            & 2,733          & 3              & 0            \\ \hline
\end{tabular}
\label{tab:attack performance}
\end{table*}

The MTP also produces a model-targeted lower bound on the size of the poisoning set required to attain the input target model. While the lower bound has no direct interpretation regarding the underlying subpopulation attack objective, it can be used as a further measure of the optimality of the MTP in the model-targeted sense. The minimum observed MTP lower bound over all attempted target models represents a lower bound on the attack difficulty, if the induced model (must) come from the set of chosen target models.

\begin{figure*}[tb]
    \centering
    \begin{subfigure}[b]{0.45\textwidth} 
        \centering
        \includegraphics[width=1.0\textwidth]{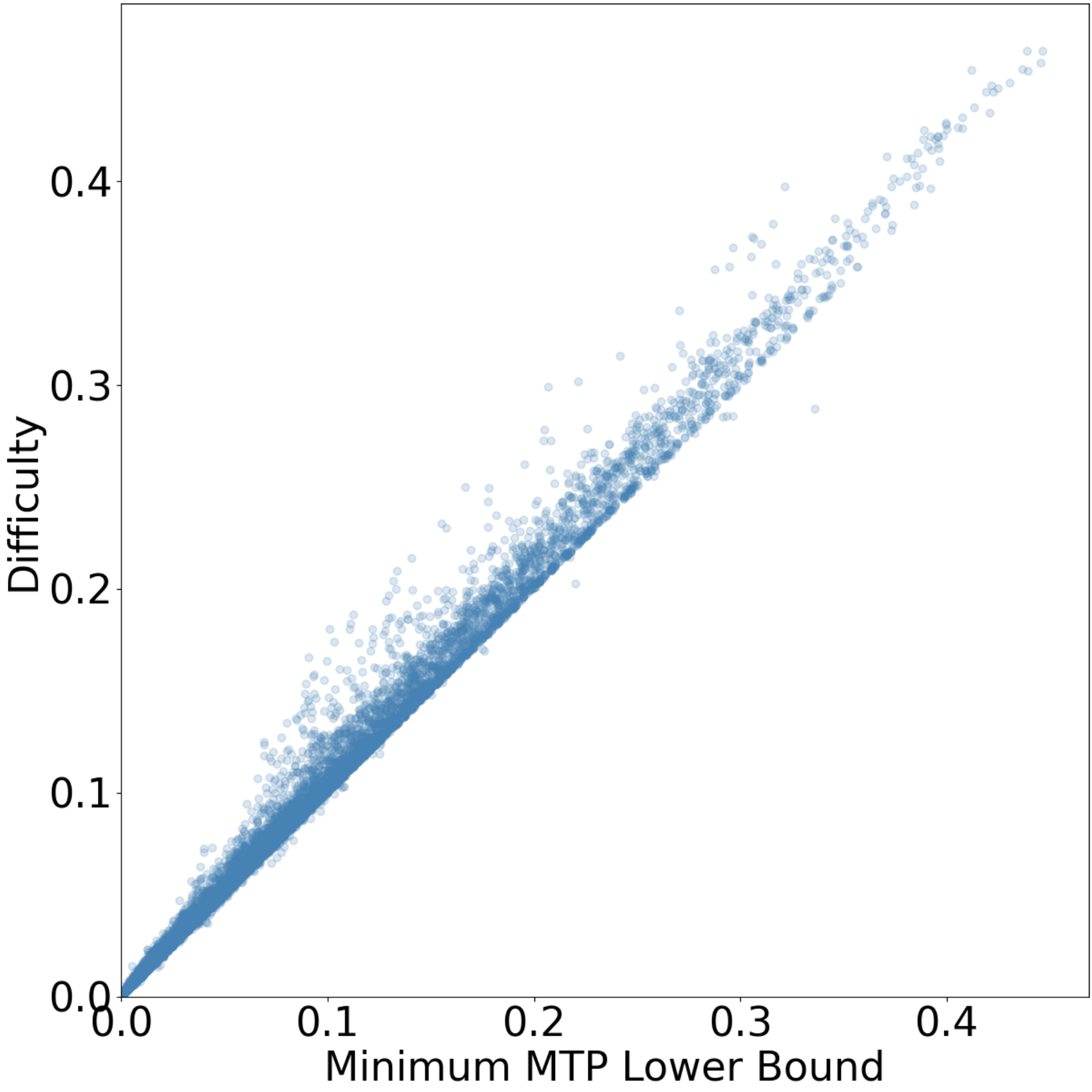}

        \caption[]%
        {Synthetic Datasets}
        \label{fig:mtp optimality synthetic}
    \end{subfigure}
    \begin{subfigure}[b]{0.45\textwidth} 
        \centering
        \includegraphics[width=1.0\textwidth]{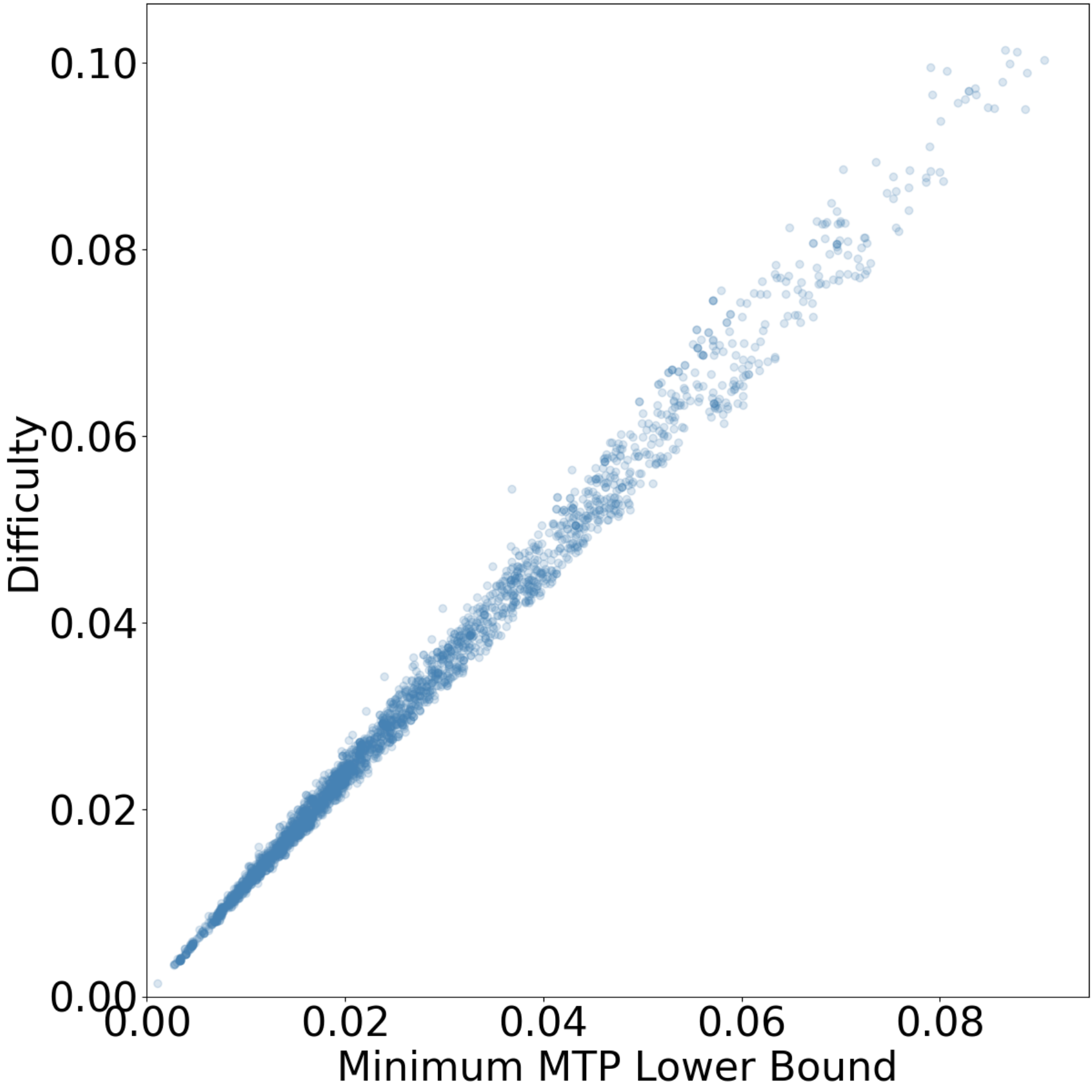}

        \caption[]%
        {Adult Dataset}    
        \label{fig:mtp optimality adult}
    \end{subfigure}
    \caption[Measuring optimality of MTP attacks]{Empirical attack difficulty versus minimum recorded model-targeted lower bound across all MTP attacks. A minimum recorded lower bound which is close to the empirical difficulty implies that there do not exist significantly more efficient attacks which exactly achieve any of the considered target models. While this does not imply any guarantee any lower bound on the number of poisoning points required to achieve the subpopulation attack objective, it does suggest that any significant differences in the true attack difficulty result from considering more optimal target models, rather than due to inefficiency of the MTP.}
    \label{fig:mtp optimality}
\end{figure*}

\Cref{fig:mtp optimality} plots the relationship between the empirical attack difficulty and the minimum recorded model-targeted lower bound among all the target models from the MTP attacks. When the observed difficulty is greater than the minimum lower bound, it indicates that there exists a target model with a potentially more effective model-targeted attack (and hence, a more effective attack against the subpopulation). Most attacks fall into this category, indicating that one of the supplied target models may admit a better model-targeted attack. However, in most cases, the bound is tight, indicating that significant improvements on attack efficiency are more likely to result from considering a larger family of target models, instead of investing in the model-targeted methodology. When the observed difficulty is less than the minimum lower bound, it indicates that the (current) most efficient attack against the subpopulation induced a model which was significantly different from any of the supplied target models. In this case, running the MTP attack an additional time using the best induced model as the target model would yield a lower bound which is smaller than the observed difficulty. However, very few attacks fall into this category, indicating that most attacks using the MTP likely succeeded by inducing a model close to one of the target models (which was potentially one of the induced models from the first round of the MTP). 

%% file: 03_synthetic_experiments.tex
\section{Synthetic Experiments}
\label{sec:subpoplulation synthetic experiment}
We provide details on the experimental setup for the synthetic experiments in \Cref{sec:subpoplulation synthetic experimet setup}. Then, we explain the dataset-level properties which contribute to attack difficulty (\Cref{sec:synthetic subpopulation variability}) and present findings on how subpopulation properties impact susceptibility to poisoning attacks (\Cref{sec:synthetic subpopulation property}).

\subsection{Experiment Setup}\label{sec:subpoplulation synthetic experimet setup}
\shortsection{Dataset Generation}
The synthetic datasets are generated using generation algorithms from Scikit-learn \citep{scikit-learn} using the techniques used to generate the ``Madelon'' dataset \citep{guyon2004nips}. The global properties of each dataset are controlled by two parameters:
\begin{enumerate}
    \item class separation, $\alpha \ge 0$ --- controls the separation between class centers. Larger values of $\alpha$ correspond to an easier classification task. 
    \item label noise parameter, $\beta \in [0, 1]$ --- controls the amount of label noise present in the data. Larger values of $\beta$ result in a less well-posed classification task, as the input features correlate less strongly with the assigned label.
\end{enumerate}
The generation process is analogous to the following procedure. First, two clusters in a 2-dimensional feature space are created by sampling from a Gaussian mixture with two components of parameters $\mathcal{N}(\gamma_1, \Sigma_1)$ and $\mathcal{N}(\gamma_2, \Sigma_2)$, respectively, with equal mixture weights. The $2\times 2$ covariances $\Sigma_1$ and $\Sigma_2$ are determined independently at random and correspond to multiplying non-translated points by a random matrix with entries sampled uniformly at random.
The distance between class centers, $\lVert \gamma_1 - \gamma_2 \rVert$, is proportional to the class separation parameter $\alpha$. The label noise parameter $\beta$ determines the fraction of points whose labels are assigned uniformly at random from $\{-1, +1\}$; the remaining labels are determined according to the corresponding component in the mixture (points sampled from $\mathcal{N}(\gamma_1, \Sigma_1)$ receive the label $-1$, and points sampled from $\mathcal{N}(\gamma_2, \Sigma_2)$ receive the label $+1$).

Datasets are generated over a grid of dataset parameters $(\alpha, \beta)$. The class separation parameter $\alpha$ ranges over thirteen equally-spaced values in the range $[0, 3]$, and the label noise parameter $\beta$ ranges over eleven equally-spaced values in the range $[0, 1]$. For each dataset parameter combination, ten datasets are generated by feeding different random seeds into the generation algorithm. Seeds are reused between different dataset parameter combinations. This results in a total of 1,430 synthetic datasets. 

 \shortsection{Model Training} 
We use linear SVM models trained using the Scikit-learn package \citep{scikit-learn} with the regularization hyperparameter $\lambda= 5 \times 10^{-4}$ for the synthetic dataset. This value was chosen to facilitate training stability across all choices of dataset parameters--earlier experiments with larger values of $\lambda$ caused instabilities during training within our learning setup when adding poisoning points to some datasets.

\shortsection{Subpopulation generation}
We identify target subpopulations following the experimental setup of Suya et al.\ \citep{suya2021model}, which is based on ClusterMatch algorithm from Jagielski et al.\ \citep{jagielski2019subpop, jagielski2021subpopulation}. We run the $k$-means clustering algorithm with $k=16$ to generate sixteen clusters, and then form target subpopulations by extracting the negative-label instances from each cluster. In total, we generate $1{,}430 \times 16 = 22{,}880$ clusters using $k$-means. From these, 21,908 non-empty subpopulations were formed from clusters containing at least one point with the negative label. Of these subpopulations, 9,591 subpopulations are trivial (i.e., the clean model already satisfies the attack objective), leaving 12,317 non-trivial subpopulations where poisoning is needed to achieve the attack goal.

The practice of attacking only the negative-labeled instances follows from prior work \citep{suya2021model} and simplifies the attack analyses by giving a clear definition of attack success as the fraction of misclassified points. While defining subpopulations in this way does not lend a natural semantic interpretation of the subpopulations, it does produce subpopulations exhibiting a wide variety of behaviors, such as size and relative position with respect to the clean model decision boundary, which are crucial for understanding the possible (inherent) factors of the subpopulation that are responsible for their disparate vulnerabilities.

\shortsection{Importance of target model selection}
Theorem 1 in Suya et al.\ \citep{suya2021model} shows that the choice of the target model plays an important role in the performance of the poisoning attacks. In particular, we desire a target model that satisfies the attack goal and also has a low loss on $\cS_c$. The drawback of choosing a poor target model (i.e., a model with higher loss on $\cS_c$) is that higher-loss target models tend to introduce additional properties irrelevant to the attack goal on the subpopulation. The MTP attack framework, which selects poisoning points based on a target model with no other knowledge of the underlying attack objective, pursues the model-targeted objective intently and consequently budgets poisoning points towards those irrelevant properties of the target model. As a result, the attack requires more poisoning points to achieve the true attack goal against the subpopulation.

\begin{figure*}[tb]
    \centering
    \begin{subfigure}[b]{0.45\textwidth}
        \centering
    \includegraphics[width=1.0\textwidth]{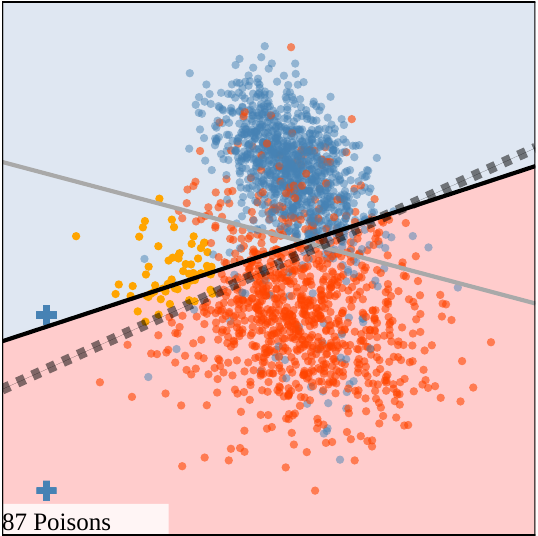}
        \caption[]%
        {Target Model with Lower Loss on $\cS_c$}
        \label{fig:target model of lower loss}
    \end{subfigure}
    \begin{subfigure}[b]{0.45\textwidth}
        \centering
    \includegraphics[width=1.0\textwidth]{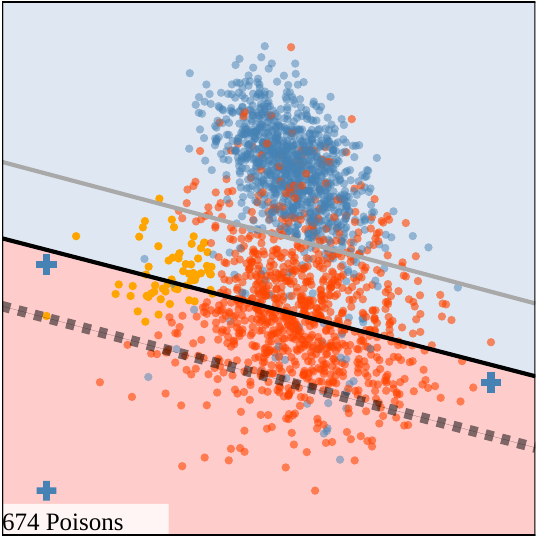}
        \caption[]%
        {Target Model with Higher Loss on $\cS_c$}
        \label{fig:target model of higher loss}
    \end{subfigure}%
    \caption[Importance of Choosing Proper Target Model for the MTP Attack]{Importance of choosing the proper target model for the MTP attack. The two figures compare the number of poisoning points needed to misclassify 50\% of the same subpopulation (orange region), but using two different target models where \Cref{fig:target model of lower loss} shows the preferred model with a lower loss on $\cS_c$. The solid gray line denotes the clean decision boundary that is trained without poisoning. The dark solid line denotes the poisoned model after adding poisoning points into the clean training set. The dashed line denotes the target model that induces 100\% test error on the selected subpopulation and is used as the target model for the MTP attack. Clean training data (identical in both figures) are depicted as points colored by label (blue for the positive label, and red for the negative label). Poisons are depicted by crosses and are colored by label.}
    \label{fig:justify target model choice}
\end{figure*}

The above argument also has a natural visual interpretation, illustrated by \autoref{fig:justify target model choice}. By pursing a target model with a higher loss on $\cS_c$, the MTP attack will experience higher resistance from the rest of the dataset (irrelevant to the attack goal on the subpopulation) as it gradually moves the decision boundary of the poisoned model in a way which compromises performance on the clean data.  \Cref{fig:target model of lower loss} shows the results of the target model that is generated by finding the model that has the lowest loss on $\cS_c$ while also satisfying the attack goals, and the MTP attack only needs 87 poisoning points to misclassify 50\% of the subpopulation. In comparison, a worse choice of target model as shown in \Cref{fig:target model of higher loss} will receive higher resistance from the rest of the poisoning points and leads to MTP using 647 poisoning points to induce 50\% of test errors on the same subpopulation. Therefore, future exploration on model-targeted attacks should also focus on finding target models that satisfy the attack objectives while having the lowest possible loss on $\cS_c$.

\begin{figure*}[tb]
    \centering
    \includegraphics[width=0.6\textwidth]{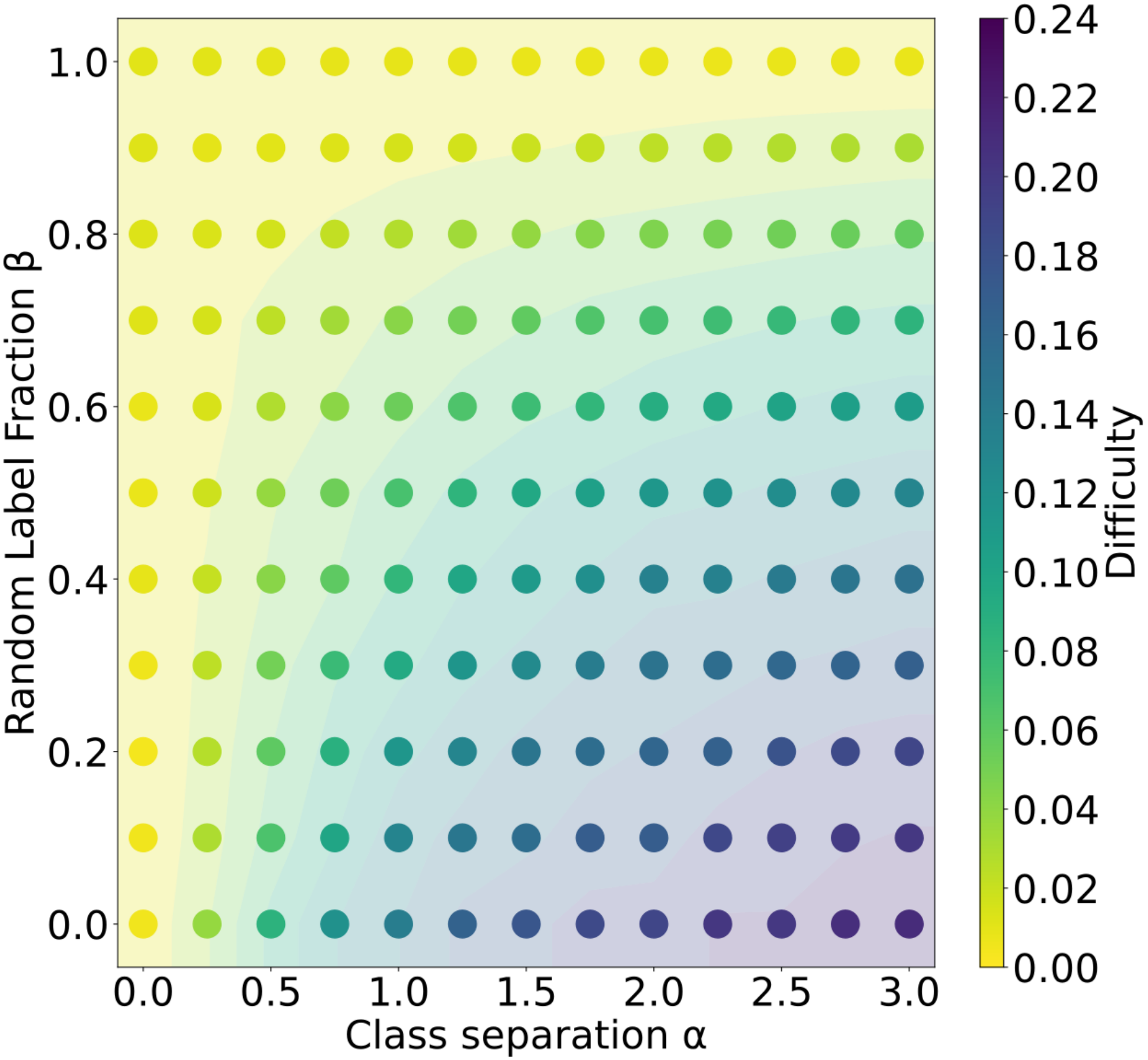}
    \caption[Comparing the Average Subpopulation Difficulty for Different Synthetic Datasets]{Comparison of the average subpopulation difficulty (over all nontrivial subpopulations) of different synthetic datasets. The \emph{difficulty} is measured as the smallest observed ratio $|\cS_p|/|\cS_c|$, where $\cS_p$ is generated to let the induced model misclassify 50\% of the points in a subpopulation.}
    \label{fig:subpopulation average difficulty}
\end{figure*}

\begin{figure*}[tb]
    \centering
    \begin{subfigure}[b]{0.45\textwidth} 
        \centering 
        \includegraphics[width=1.0\textwidth]{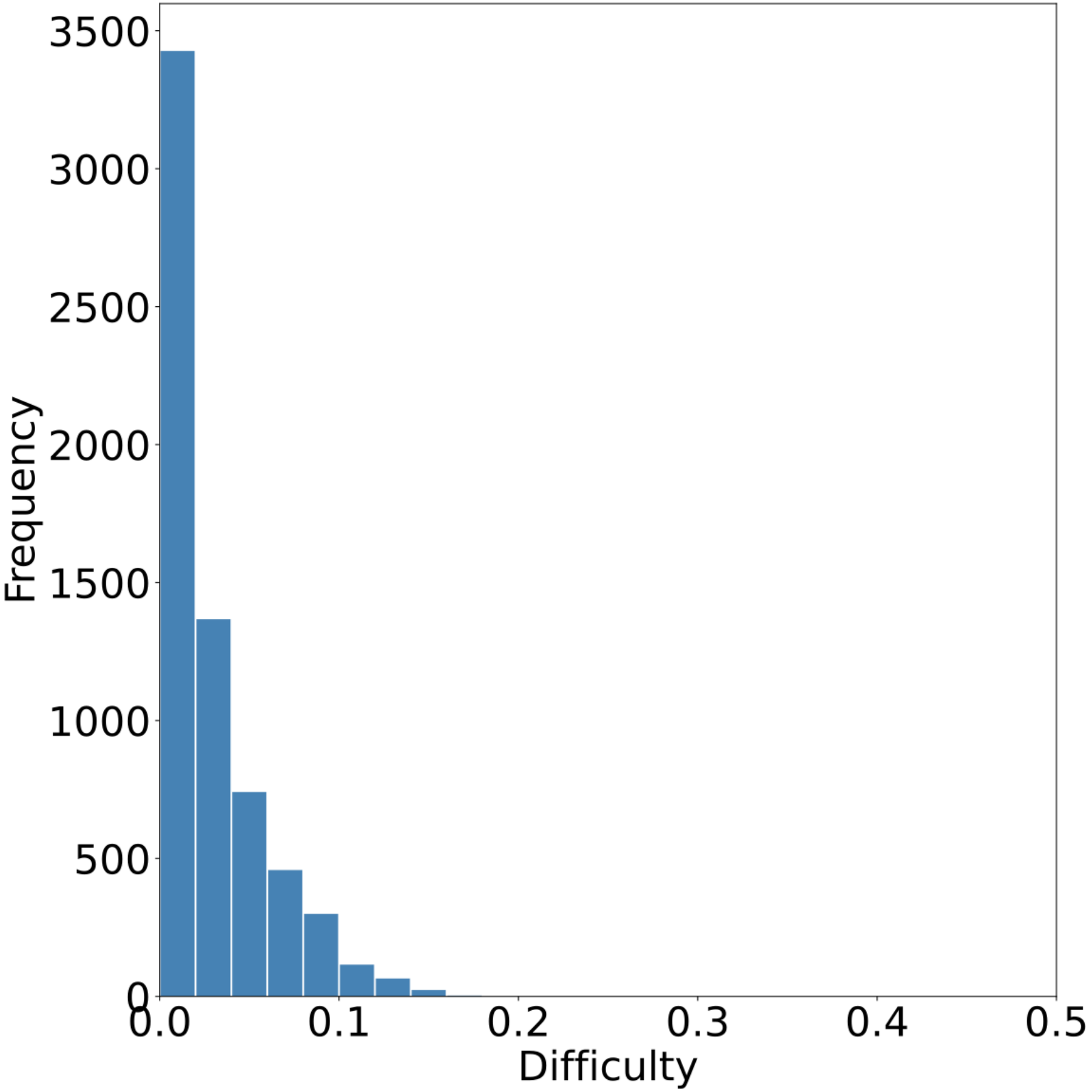}

        \caption[]%
        {Inaccurate Clean Models}    
        \label{fig:difficulty for inaccurate clean models}
    \end{subfigure}
    \begin{subfigure}[b]{0.45\textwidth} 
        \centering 
        \includegraphics[width=1.0\textwidth]{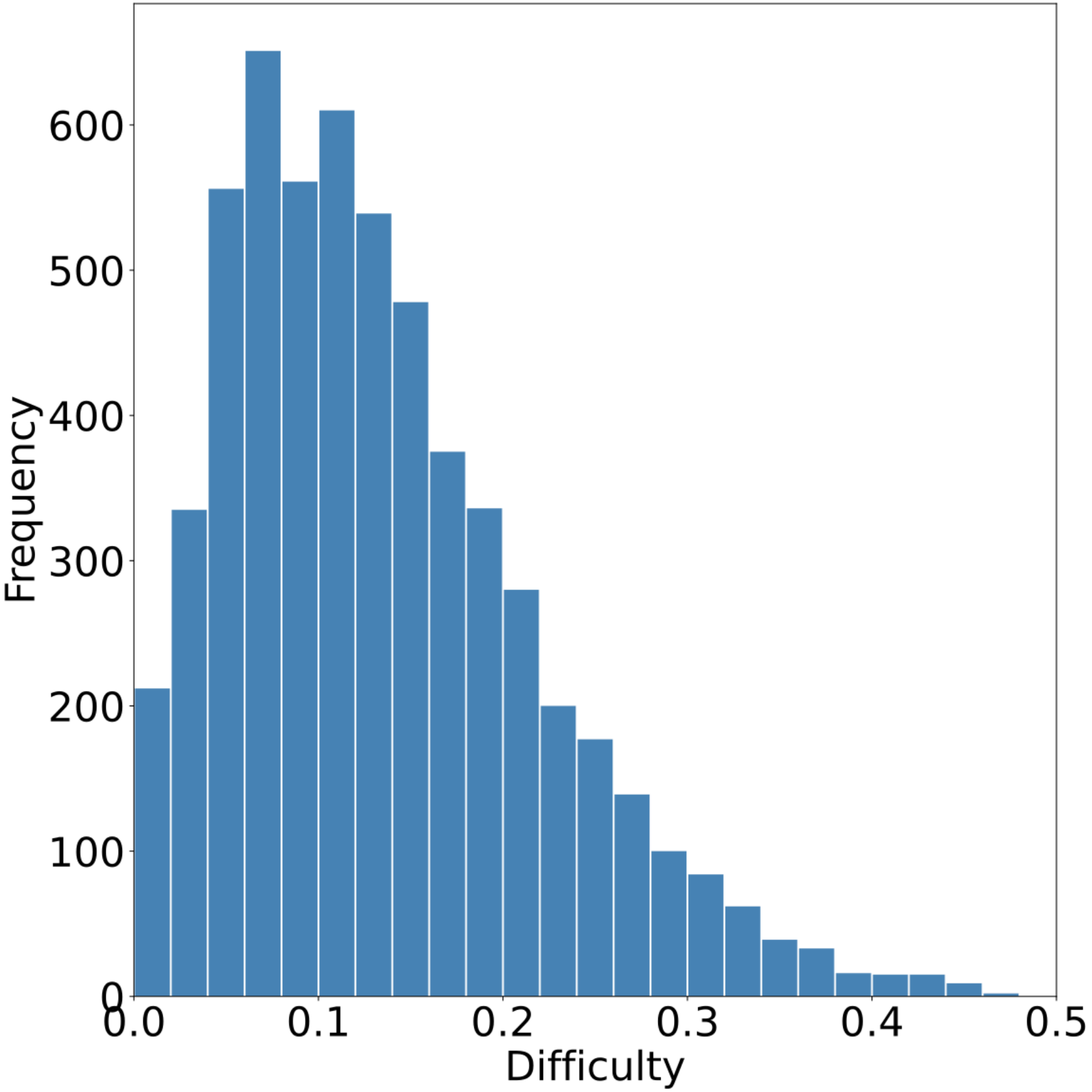}
        
        \caption[]%
        {Accurate Clean Models}    
        \label{fig:difficulty for accurate clean models}
    \end{subfigure}
    \caption[Distribution of Subpopulation Susceptibilities for Synthetic Datasets]
    {Distribution of subpopulation susceptibilities for synthetic datasets of different clean accuracies under linear SVM. The x-axis denotes the measured attack difficulty and the y-axis denotes the frequency. \Cref{fig:difficulty for inaccurate clean models} depicts the attack difficulty distribution for inaccurate clean models ($\le 70\%$ accuracy on clean dataset). \Cref{fig:difficulty for accurate clean models} depicts the attack difficulty distribution for accurate clean models ($> 70\%$ accuracy on clean dataset).
    }
    \label{fig: susceptibility variation frequencies}
\end{figure*}

\subsection{Subpopulation Susceptibility Variation}\label{sec:synthetic subpopulation variability}
In this section, we describe the variation in subpopulation susceptibility by first showing the (dataset-level) average difficulty of subpopulations for each dataset and then providing a finer analysis of the variation across individual subpopulations for different datasets. 

\shortsection{Average subpopulation susceptibility across datasets}
We first explore the impact of the overall distributional properties on the vulnerabilities of the subpopulations, as these are high-level properties that might provide some general insights before digging deeper into particular subpopulation properties. \autoref{fig:subpopulation average difficulty} shows the average attack difficulty as a function of the dataset class separation parameter $\alpha$ and the label noise parameter $\beta$, where the average is taken over all nontrivial subpopulations (i.e., at least one poisoning point is required to achieve the attack objective) and over all 10 dataset seeds. For datasets that are easier to classify (e.g., higher class separation $\alpha$ and lower fraction of label noise $\beta$ from the distributional perspective), the average difficulty of the subpopulations increases, and vice versa. This observation is expected as the poorly separated datasets either already have high test errors on the subpopulations without poisoning or have points that are close to the decision boundary and are highly sensitive to misclassification when the decision boundary changes slightly, indicating that higher test errors on the subpopulations after poisoning can be easily achieved with a limited number of poisoning points. In contrast, for datasets that are well-separated by linear models, the subpopulations are far from the decision boundary with a lower number of misclassified points, and more poisoning points are needed to move the decision boundary (significantly) to incur the desired amount of test errors on the subpopulations. To conclude, the overall distributional properties of class separation and label noise indeed have a major impact on the vulnerabilities of the subpopulations, and subpopulations in less separable datasets are more vulnerable to poisoning.

\shortsection{Distribution of subpopulation susceptibilities} 
Once we have an understanding of the average difficulty for the subpopulations in each dataset, we further explore the variation of subpopulation susceptibility (measured by the attack difficulty) across different subpopulations for both the poorly-separated and well-separated datasets. We plot the frequencies of subpopulations with respect to their difficulty scores in \autoref{fig: susceptibility variation frequencies}. From these experiments, we find that when the clean model accuracy is low (i.e., datasets are less separable under linear SVM), most subpopulations are easy to attack. In these cases, the distributional properties dominate the subpopulation susceptibility, and attack difficulty exhibits limited dependence on individual subpopulation properties. 
However, when the overall clean accuracy is high, the variation of the difficulty becomes more drastic, and the impact of individual subpopulations matters more for the susceptibility. %

\begin{figure*}[tb]
    \centering
    \begin{subfigure}[b]{0.45\textwidth} 
        \centering 
        \includegraphics[width=1.0\textwidth]{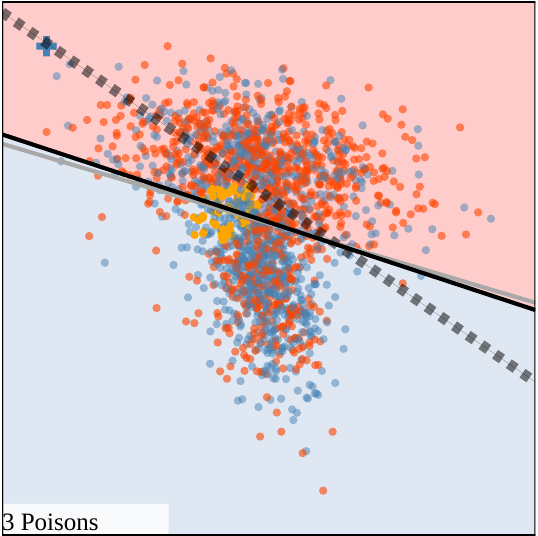}
        \caption[]%
        {Easiest to Attack}    
    \end{subfigure}
    \begin{subfigure}[b]{0.45\textwidth} 
        \centering 
    \includegraphics[width=1.0\textwidth]{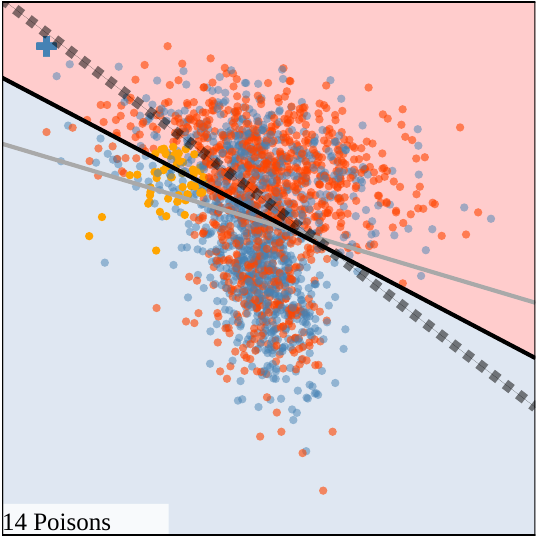}
        \caption[]%
        {Moderately Hard to Attack}    
    \end{subfigure}
    \begin{subfigure}[b]{0.45\textwidth} 
        \centering 
    \includegraphics[width=1.0\textwidth]{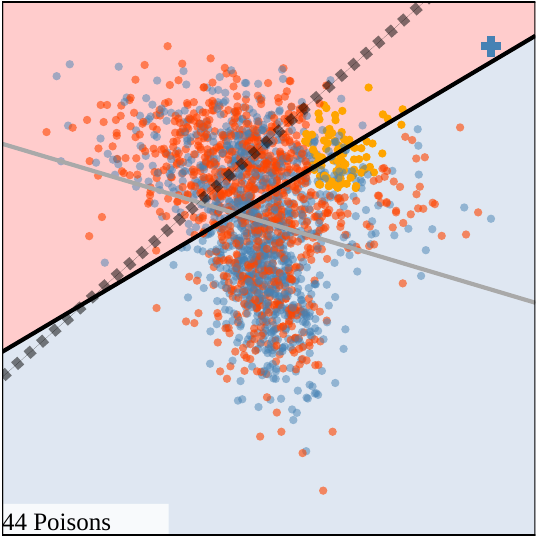}
        \caption[]%
        {Moderately Hard to Attack}    
    \end{subfigure}
    \begin{subfigure}[b]{0.45\textwidth} 
        \centering 
    \includegraphics[width=1.0\textwidth]{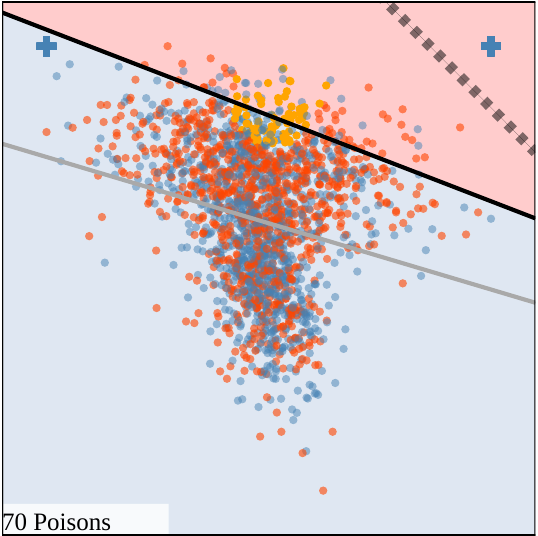}
        \caption[]%
        {Hardest to Attack}    
    \end{subfigure}
    \caption[Variation of Subpopulation Susceptibilities in the Non-linearly Separable Synthetic Dataset]
    {Variation of subpopulation susceptibilities in a non-linearly separable synthetic dataset. Each figure contains the number of poisoning points used by the most efficient attack from our experiments to have the induced model misclassify (as blue) 50\% of the points in the subpopulation (colored in orange). The solid gray line denotes the clean decision boundary that is trained without poisoning. The dark solid line denotes the poisoned model after adding poisoning points into the clean training set. The dashed line denotes the target model that was used for the attack.
    }
    \label{fig: susceptibility variation non-separable dataset}
\end{figure*}

\begin{figure*}[t!]
    \centering
    \begin{subfigure}[b]{0.45\textwidth} 
        \centering 
        \includegraphics[width=1.0\textwidth]{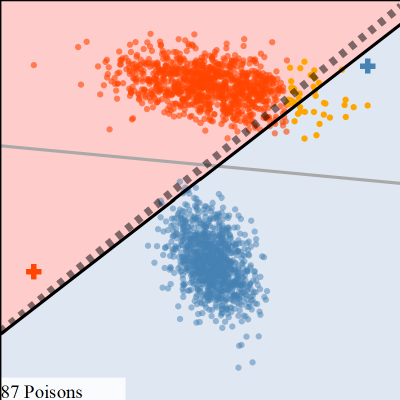}
        \caption[]%
        {Easiest to Attack}    
        \label{fig: susceptibility variation separable dataset easy}
    \end{subfigure}
    \begin{subfigure}[b]{0.45\textwidth} 
        \centering 
        \includegraphics[width=1.0\textwidth]{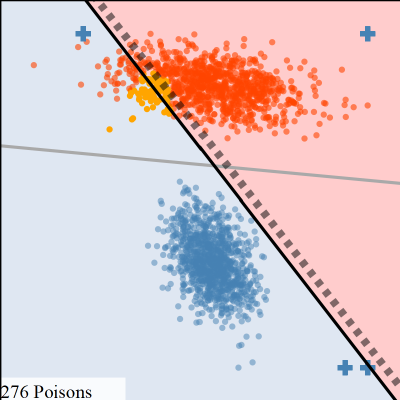}
        \caption[]%
        {Moderately Hard to Attack}    
        \label{fig: susceptibility variation separable dataset mid1}
    \end{subfigure}
    \begin{subfigure}[b]{0.45\textwidth} 
        \centering 
    \includegraphics[width=1.0\textwidth]{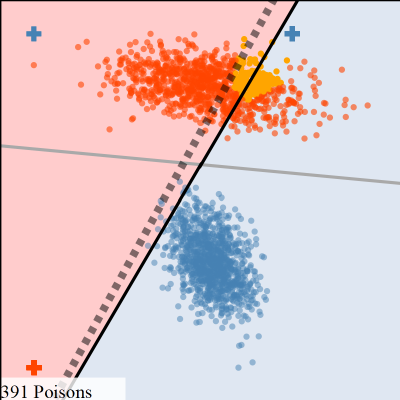}
        \caption[]%
        {Moderately Hard to Attack}    
        \label{fig: susceptibility variation separable dataset mid2}
    \end{subfigure}
    \begin{subfigure}[b]{0.45\textwidth} 
        \centering 
    \includegraphics[width=1.0\textwidth]{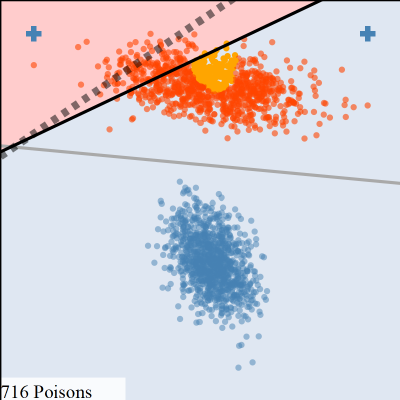}
        \caption[]%
        {Hardest to Attack}    
        \label{fig: susceptibility variation separable dataset hard}
    \end{subfigure}
    \caption[Variation of Subpopulation Susceptibilities in the Linearly Separable Synthetic Dataset]
    {Variation of subpopulation susceptibilities in the linearly separable synthetic dataset. Each figure contains the number of poisoning points used by the most efficient attack from our experiments to have the induced model misclassify (as blue) 50\% of the points in the subpopulation (colored in orange). The solid gray line denotes the clean decision boundary that is trained without poisoning. The dark solid line denotes the poisoned model after adding poisoning points into the clean training set. The dashed line denotes the target model that was used for the attack.
    }
    \label{fig: susceptibility variation separable dataset}
\end{figure*}

\shortsection{Understanding subpopulation susceptibilities through visualization}  \Cref{fig: susceptibility variation non-separable dataset} shows 4 different subpopulations that include the hardest, easiest, and moderately hard to attack subpopulations for a poorly-separated dataset under linear SVM. We find that in general, subpopulations in less separable datasets are more vulnerable to poisoning: the most expensive attack depicted in \Cref{fig: susceptibility variation non-separable dataset} requires a poisoning fraction of only $70/2000 = 3.5\%$. However, a notable finding is that individual subpopulation properties still lead to some minor variation in their susceptibilities to poisoning. Expectedly, this variation across subpopulations is amplified for well-separated datasets. \Cref{fig: susceptibility variation separable dataset} shows 4 different subpopulations (hardest, easiest, and moderately hard to attack) in the well-separated case, where the difficulties of the depicted attacks range from just over 5\% of the training set size to just under 36\%. Finally, the visualizations in \Cref{fig: susceptibility variation non-separable dataset} and \Cref{fig: susceptibility variation separable dataset} suggest an informal geometric property which may affect attack difficulty: the relative position of the subpopulation to the rest of the population. We attempt to concretely measure the informal notion of ``relative position" in \Cref{sec:synthetic subpopulation property}.

\begin{figure*}[t!]
    \centering
    \begin{subfigure}[b]{0.45\textwidth} 
        \centering 
        \includegraphics[width=1.0\textwidth]{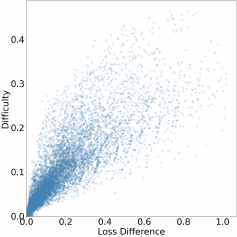}
        
        \caption[]%
        {Model Loss Difference}   
        \label{fig: synthetic subpop factors loss diff}
    \end{subfigure}
    \begin{subfigure}[b]{0.45\textwidth} 
        \centering 
    \includegraphics[width=1.0\textwidth]{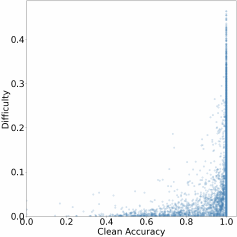}
        \caption[]%
        {Accuracy on Clean Subpopulation Points}    
        \label{fig: synthetic subpop factors subpop acc}
    \end{subfigure}
    \\[3ex] 
    \begin{subfigure}[b]{0.45\textwidth} 
        \centering 
    \includegraphics[width=1.0\textwidth]{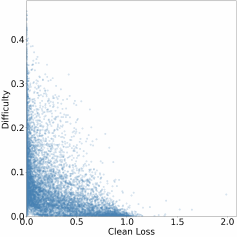}
        
        \caption[]%
        {Loss on Clean Subpopulation Points}    
        \label{fig: synthetic subpop factors subpop loss}
    \end{subfigure}
    \begin{subfigure}[b]{0.45\textwidth} 
        \centering 
    \includegraphics[width=1.0\textwidth]{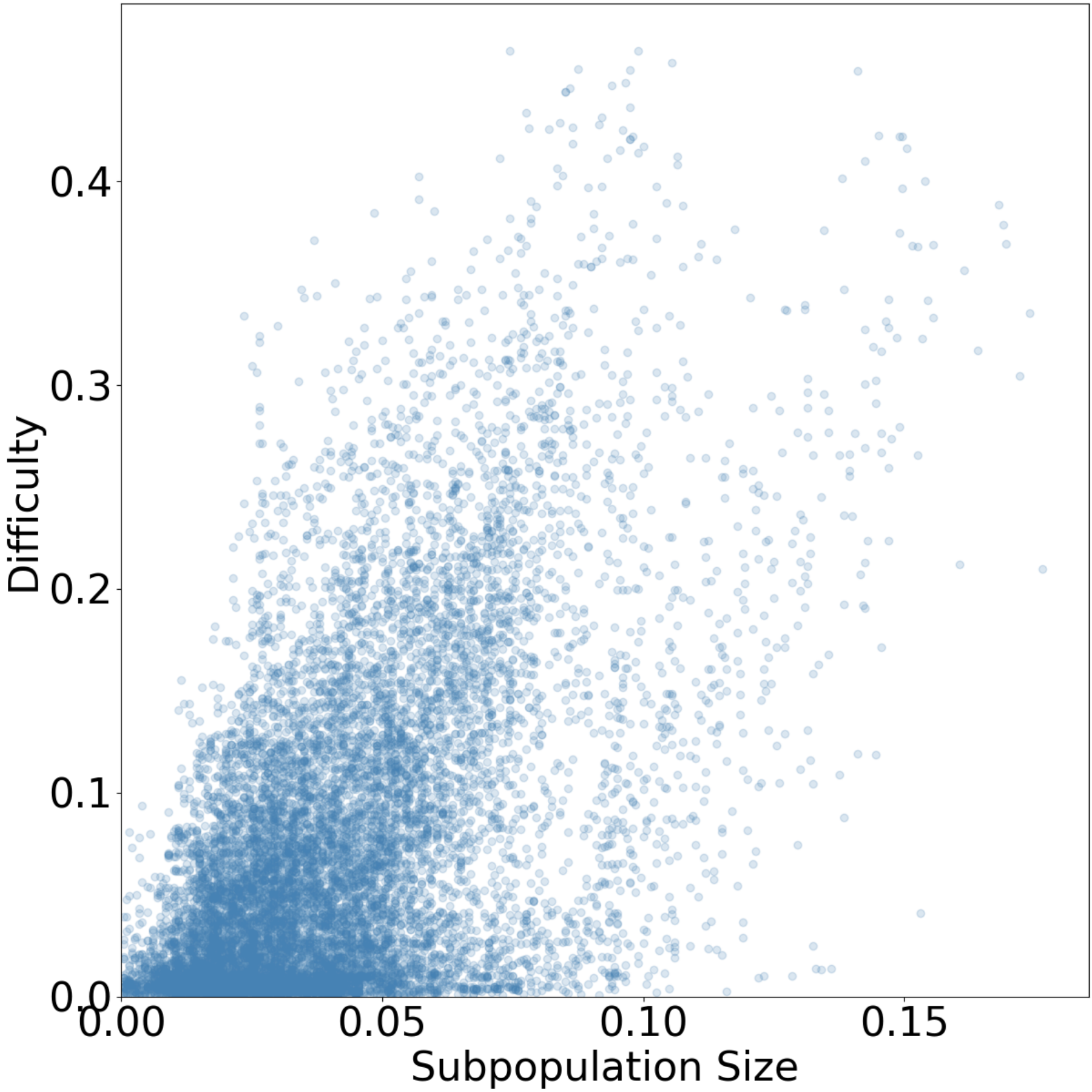}

        \caption[]%
        {Subpopulation Size} 
        \label{fig: synthetic subpop factors subpop size}
    \end{subfigure}
    \caption[Synthetic Dataset: Correlation Between Subpopulation Properties and Susceptibilities]
    {Correlation between the properties of the subpopulation and the susceptibilities for the synthetic dataset. \Cref{fig: synthetic subpop factors loss diff} shows the correlation of the model loss difference between the generated target model (has $\ge$ 50\% error on the subpopulation and also the lowest loss on $\cS_c$ among the generated candidate target models) and the clean model on $\cS_c$. \Cref{fig: synthetic subpop factors subpop acc} shows the correlation of the subpopulation clean accuracy. \Cref{fig: synthetic subpop factors subpop loss} shows the correlation of the clean loss on the subpopulation. \Cref{fig: synthetic subpop factors subpop size} shows the correlation of the subpopulation size. 
    }
    \label{fig: synthetic subpop factors}
\end{figure*}

\subsection{Characterization of Subpopulation Properties Impacting  Susceptibility}\label{sec:synthetic subpopulation property}
Through our experiments we have demonstrated that, broadly speaking, subpopulation poisoning attack difficulty is primarily determined by distributional properties when sampled datasets are poorly separated by linear models (while individual subpopulations still contribute to minor variations). But when datasets are well-separated, subpopulation properties that capture the ``relative positions" of the subpopulations contribute more significantly to attack difficulty. In this section, we attempt to identify the relevant subpopulation properties which correspond to this geometric notion. We test four factors that are explicitly related to the properties of the subpopulations and also the underlying model (i.e., linear SVM in our case): 1) the minimum observed model loss difference between a target model and the clean model on $\cS_c$, where the considered target models are those generated from the label-flipping attack and achieve at least 50\% error on the subpopulation; 2) the training accuracy of the clean model on the subpopulation; 3) the training loss of the clean model on the subpopulation; and 4) the size of the subpopulation as a fraction of the clean training set size. 

\Cref{fig: synthetic subpop factors} shows the correlation between the four factors and the subpopulation susceptibility. Only the model loss difference appears to have a strong correlation with the empirically observed susceptibility using our attacks, while the other factors of clean accuracy and clean loss on subpopulation and the size of the subpopulation do not have a significant correlation by themselves.

We believe the model loss difference is a reliable indicator of difficulty because it implicitly captures the relative position of the subpopulation to the rest of the population by examining the space of target models which achieve the attack objective (i.e., misclassify the subpopulation). The existence of a target model with a small loss difference to the clean model suggests that 
predictions on the subpopulation can be made independently of predictions on the rest of the population. In this case, poisoning is likely to be effective because the points outside the subpopulation (which affect the resulting model via the loss function) do not need to interfere with the goals of the adversary. When such a target model does not exist (i.e., all target models which achieve the attack objective also incur a large error on the larger population), it is likely that poisoning attacks against the subpopulation will be less effective. Stated another way, attacks are more efficient when the adversary is able to move the decision boundary of the poisoned model without facing resistance from the clean data. 

Other factors such as the clean accuracy and the clean loss are related to the average margin of the clean points in the subpopulation to the decision boundary, but do not capture information about the relationship between the subpopulation and the rest of the clean data. If the subpopulation lies close to the decision boundary, but is surrounded in all directions by the rest of the clean data which the clean model can confidently separate, then misclassifying the subpopulation will unavoidably misclassify points from the rest of the population and thus require more poisoning points to counter the stronger resistance from clean data. Subpopulation size similarly fails to characterize the susceptibility, since the metric does not on its own capture information about the spacial relationship between the subpopulation and the rest of the subpopulation.

The above argument is also visually supported by \autoref{fig: susceptibility variation separable dataset}, which demonstrates several different subpopulations in the same dataset which appear similar as measured by clean model behavior and subpopulation size, yet exhibit significantly different empirical attack difficulties. In particular, the subpopulations in \Cref{fig: susceptibility variation separable dataset easy} and \Cref{fig: susceptibility variation separable dataset mid1} are both classified with perfect accuracy by the clean model, sit at roughly the same distance from the decision boundary and are of similar sizes. Yet, empirical attacks against the subpopulation in \Cref{fig: susceptibility variation separable dataset easy} require only half as many poisoning points as attacks against the subpopulation in \Cref{fig: susceptibility variation separable dataset mid1}, in part due to the geometric properties of the data and the space of target models that satisfy either attack objective.

In these experiments, we only measure the correlation of the factors individually and observe the latter three factors are not relevant. However, using more sophisticated ways to combine these irrelevant factors may still have strong predictive power on the subpopulation susceptibility. In addition, these irrelevant factors may still become relevant when we limit our scope to understanding the vulnerabilities of selected subpopulations. We leave these explorations as future work.

%% file: 04_adult_experiments.tex
\section{Experiments on Adult Dataset}
\label{sec:subpoplulation adult experiment}

In addition to motivating the factors affecting subpopulation vulnerability using low-dimensional synthetic datasets, we show that variability in subpopulation vulnerability arises in a complex, practical setting. Following the initial observations by Jagielski et al.~\citep{jagielski2019subpop, jagielski2021subpopulation}, we conduct poisoning attacks against a large number of subpopulations in the UCI Adult dataset.

We show the experimental setup for the Adult dataset in \Cref{sec:subpoplulation adult experimet setup}. Then we show the generally descriptive subpopulation properties that are related to the subpopulation susceptibility in \Cref{sec:subpoplulation adult experimet results} and present some semantically meaningful subpopulation properties that are related to the susceptibility in selected settings in \Cref{sec:subpoplulation adult experimet semantic}.

\subsection{Experiment Setup}
\label{sec:subpoplulation adult experimet setup}

\shortsection{Benchmark Dataset and Model}
We use the UCI Adult dataset \citep{Dua:2019}, which has been used in prior work on subpopulation poisoning attacks \citep{jagielski2019subpop, jagielski2021subpopulation, suya2021model}, as the tabular benchmark dataset to validate our findings in a more complex and practical setting. The learning task is to predict whether a person's income is greater than \$50K per year using demographic information including age, race, sex, and level of education gathered from US census data. We adopt the data preprocessing steps from prior work \citep{suya2021model}, resulting in a class-balanced, downsampled dataset of 15,682 training points and 7,692 test samples. Each example is 57-dimensional after applying one-hot encoding to categorical attributes. As with the synthetic data, we again use linear SVM models trained using the Scikit-learn package with the regularization hyperparameter $\lambda = 0.09$ for the Adult dataset.

\shortsection{Subpopulation Generation}
We generate subpopulation for the Adult dataset using a technique based on the FeatureMatch \citep{jagielski2019subpop, jagielski2021subpopulation} algorithm, which filters data according to a set of pre-selected attributes to produce a subpopulation. Subpopulations generated this way are more semantically meaningful since they correspond to realistic demographic groups (e.g., middle-aged service workers or women with college degrees) and are related to established definitions of group fairness in the field of algorithmic fairness \citep{buolamwini2018gender, hardt2016equality, dwork2012fairness}. These subpopulations also coincide with the type of subpopulations an adversary might target in practice, since a realistic adversary is likely to be concerned with controlling model behavior on data inputs with semantically meaningful characteristics.

To generate each subpopulation, we first select a subset of categorical features and choose specific values for those features. For example, the categorical features could be chosen to be ``work class'' and ``education level'', and the features' values could then be chosen to be ``never worked'' and ``some college'', respectively. Then, every negative label (``$\leq 50\mathrm{K}$'') instance in the training set matching all the (feature, label) pairs is extracted to form the subpopulation. The subpopulations for our experiments are chosen by considering every subset of categorical features and every combination of those features that is present in both the training set and the test set. For simplicity, we only consider subpopulations with a maximum of three feature selections.

This method results in 4,338 subpopulations, each of which is targeted in a poisoning attack using the same attack as in the case of the synthetic dataset. Of these attacks, 1,602 are trivial (i.e., the clean model already satisfies the attacker objective), leaving 2,736 nontrivial attacks.

\begin{figure*}[tb]
    \centering
    \includegraphics[width=0.4\textwidth]{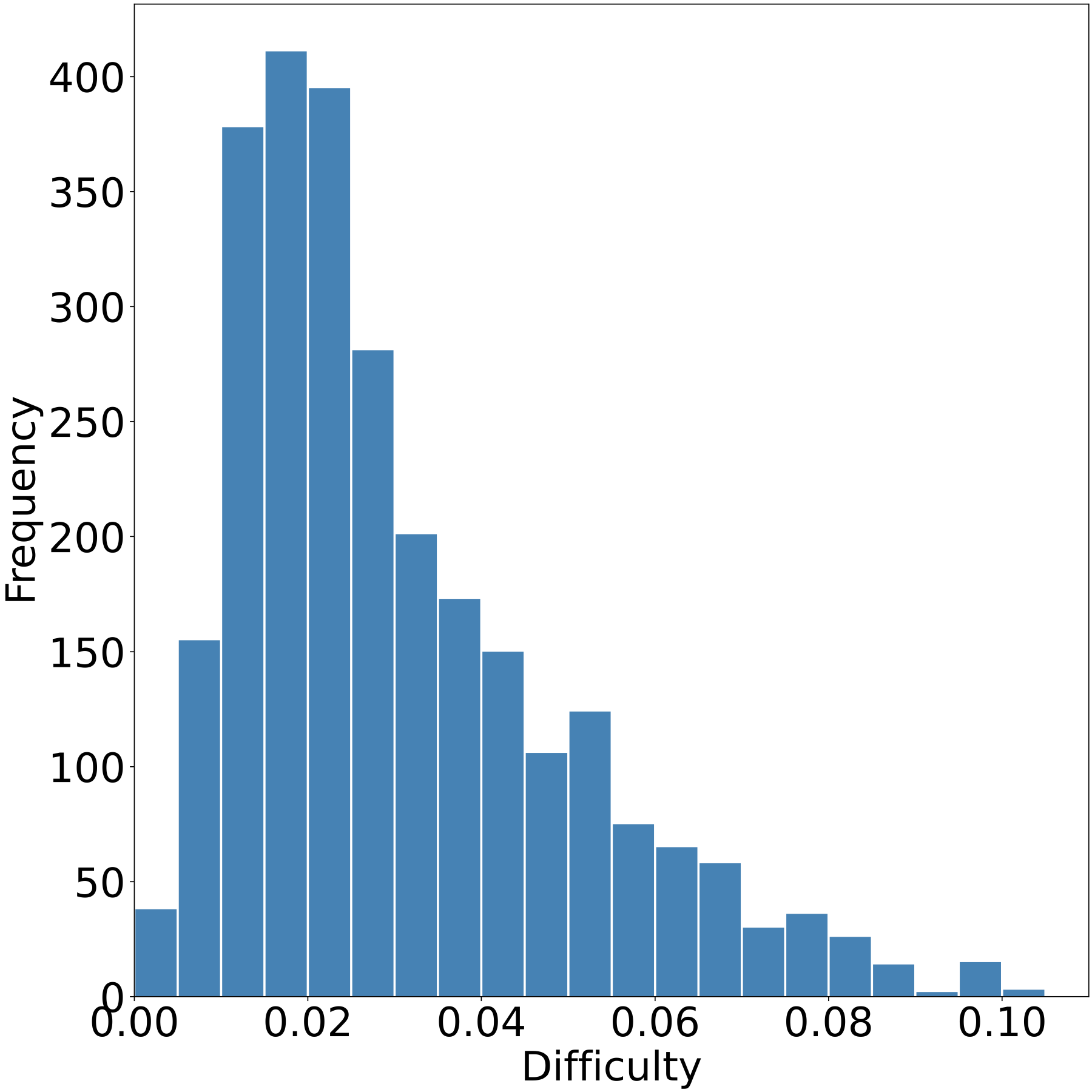}

    \caption[Variation of Subpopulation Susceptibility in the Adult Dataset]{Distribution of the attack difficulty for subpopulations of the Adult dataset.}
    \label{fig:adult subpopulation susceptibility variation}
\end{figure*}

\begin{figure*}[tb]
    \centering
    \begin{subfigure}[b]{0.45\textwidth} 
        \centering
        \includegraphics[width=1.0\textwidth]{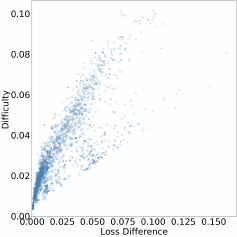}
        
        \caption[]%
        {Loss Difference between Clean / Target Models}    
    \end{subfigure}
    \begin{subfigure}[b]{0.45\textwidth} 
        \centering 
    \includegraphics[width=1.0\textwidth]{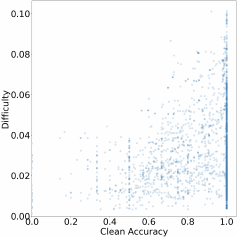}
        \caption[]%
        {Accuracy on Clean Subpopulation Points}    
    \end{subfigure}

\ \\[3ex]
    
    \begin{subfigure}[b]{0.45\textwidth} 
        \centering 
    \includegraphics[width=1.0\textwidth]{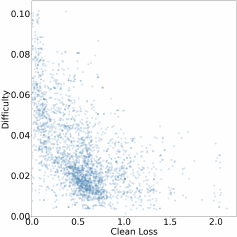}
        
        \caption[]%
        {Loss on Clean Subpopulation Points}    
    \end{subfigure}
    \begin{subfigure}[b]{0.45\textwidth} 
        \centering 
    \includegraphics[width=1.0\textwidth]{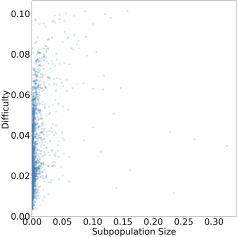}

        \caption[]%
        {Subpopulation Size}    
    \end{subfigure}
    
    \caption[Adult Dataset: Correlation between Subpopulation Properties and Susceptibilities]
    {Correlation between the properties of the Subpopulation and the susceptibilities for the Adult dataset. Model loss difference still denotes the smallest observed loss difference between a generated target model ($\ge$ 50\% error on the subpopulation with the lowest loss on $\cS_c$) and the clean model on $\cS_c$. 
    }
    \label{fig: adult subpop factors}
\end{figure*}

\subsection{Variation of Subpopulation Susceptibility and Relevant Properties}
\label{sec:subpoplulation adult experimet results}
The subpopulation susceptibility for the Adult dataset subpopulations varies drastically, as shown in \autoref{fig:adult subpopulation susceptibility variation}. %
Similarly to the synthetic case studied in \autoref{sec:subpoplulation synthetic experiment}, we proceed to explore the correlation of the four factors (i.e., model loss difference, accuracy on subpopulation, loss on subpopulation, and subpopulation size) to the subpopulation susceptibility in the Adult dataset. From \Cref{fig: adult subpop factors} we see that the model loss difference is a generalizable property that is highly correlated to the susceptibility while other factors again fail to show a strong correlation for the Adult dataset. We believe the underlying reason for this observation is similar to the synthetic case---the model loss difference well captures the relative position of the subpopulation but the other properties do not. 

\begin{figure*}[t!]
    \centering
    \includegraphics[width=0.8\textwidth]{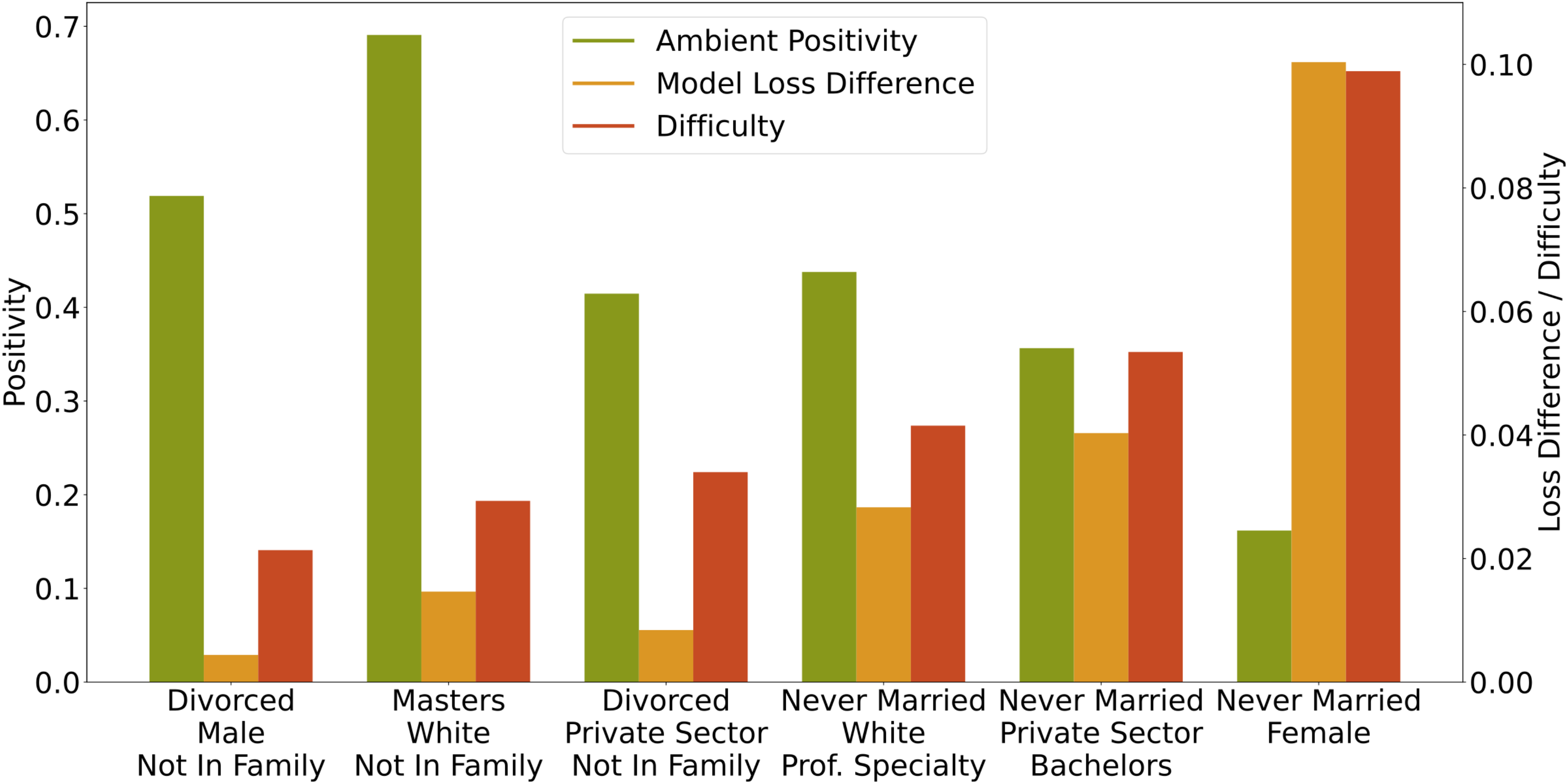}
    \caption[Adult Dataset: Correlation between Ambient Positivity and Subpopulation Susceptibility on Selected Subpopulations]{The correlation of ambient positivity on subpopulation susceptibility for selected subpopulations in the Adult dataset}
    \label{fig:adult ambient positivity}
\end{figure*}

\subsection{Related Semantic Properties}
\label{sec:subpoplulation adult experimet semantic}

Recall that for the Adult dataset, the subpopulations are generated using the FeatureMatch algorithm so each subpopulation has a semantic meaning. Hence, we can explore if there are identifiable semantic properties, rather than just the abstract descriptions using the model loss difference, that can be related to the subpopulation susceptibility. %

Through experiments, we find that the number of nearby points with different labels to the subpopulations can also be related to the subpopulation susceptibility. To show this, we first define \emph{ambient positivity}. For a subpopulation under binary classification with a given property $P$, we define the \emph{ambient subpopulation} as the set of all points satisfying $P$ (since it also includes positive-label points), and call the fraction of points in the ambient subpopulation with a positive label the \emph{ambient positivity} of the subpopulation. 

The relation between the ambient positivity and the subpopulation susceptibility is shown in \Cref{fig:adult ambient positivity} for some subpopulations. These subpopulations are chosen to have 100\% test accuracy by the clean model and are of similar sizes (ranging from 1\% to 2\% of the clean training set size $|\cS_c|$). In the above attacks, attack difficulty is negatively correlated with the ambient positivity of the subpopulation. This makes sense since positive-label points near the subpopulation work to the advantage of the attacker when attempting to induce misclassification of the negative-label points (i.e., less resistance from the rest of the population). In terms of the model-targeted attack, if the clean model classifies the ambient subpopulation as the negative label then the loss difference between the target and clean models is smaller if there are positive-label points in that region. We did not test the ambient positivity for the synthetic case because ambient positivity is highly correlated with the label noise parameter $\beta$, diminishing its relevance relative to other factors affecting subpopulation susceptibility.

\begin{figure*}[t]
    \centering
    \includegraphics[width=0.6\textwidth]{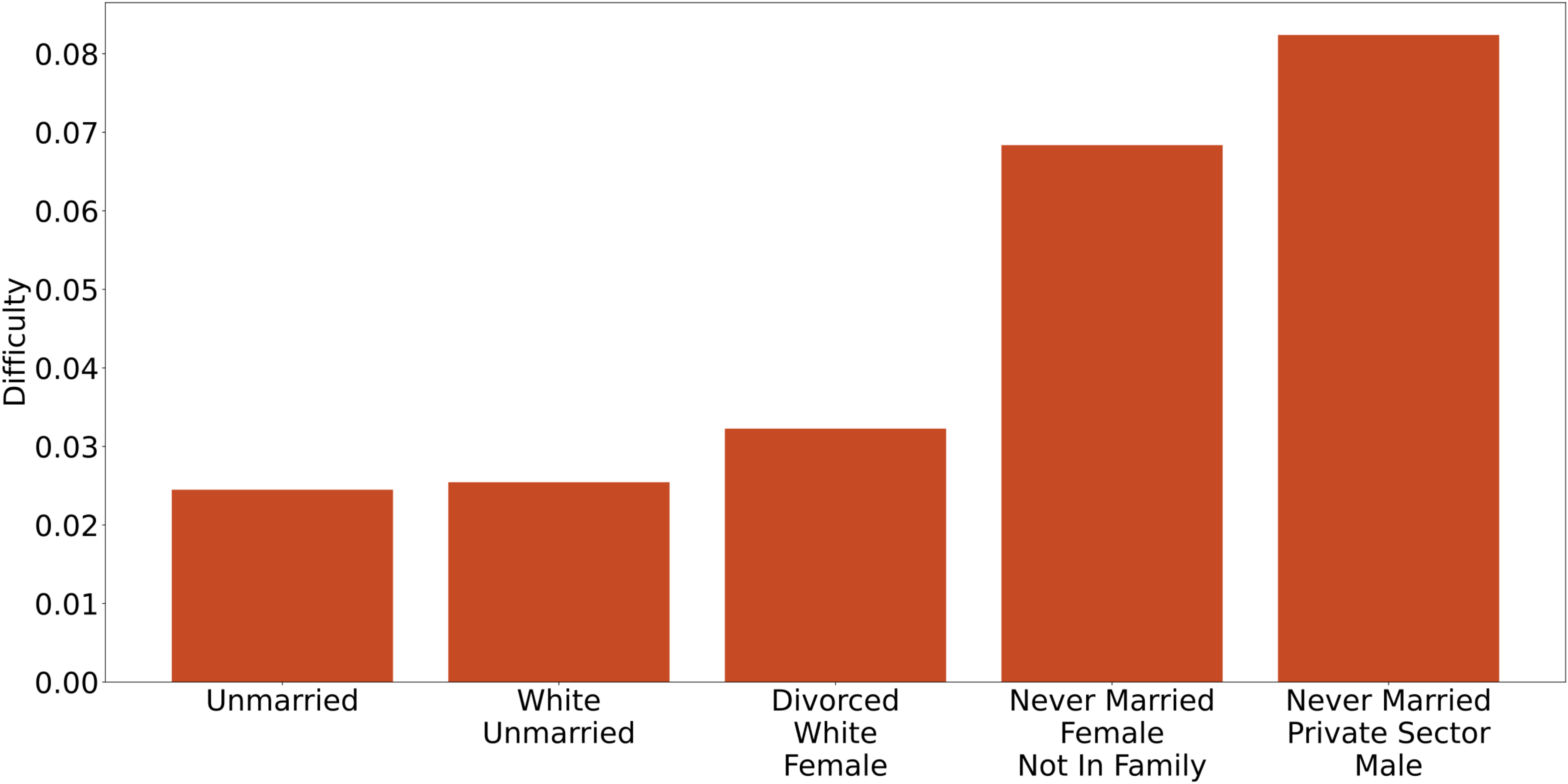}
    \caption[Adult Dataset: Limitation of Ambient Positivity on Predicting Subpopulation Susceptibility]{Variation of subpopulation susceptibility for subpopulations with similar ambient positivity in the Adult dataset.}
    \label{fig:adult ambient positivity limitation}
\end{figure*}

Next we consider if the ambient positivity of a subpopulation necessarily determines attack difficulty for otherwise similar subpopulations. We find that this is not the case---if we restrict our view to subpopulations with similar pre-poisoning ambient positivity (e.g., between 0.2 and 0.3), while still having 100\% classification accuracy and similar subpopulation size, we still find a significant spread of attack difficulties as shown in \autoref{fig:adult ambient positivity limitation}. This observation highlights the challenges in identifying generally-applicable semantic properties for explaining variation in subpopulation susceptibility. 

Related to the challenge of identifying related semantic properties, there are also subpopulations of different susceptibility that match on the same features but differ in the value of only a single feature. In addition, these subpopulations are similarly sized and also are perfectly classified by the clean model. For example, the two subpopulations that take similar values of ``Clerical'' for the feature ``Occupation'' and ``Female'' for the feature ``Sex'' only differ in the value of the feature ``Relationship Status'', and yet the one with the value ``Not In Family'' has an attack difficulty score of 0.07 while the other one with ``Unmarried'' has a difficulty score of only 0.02.

%% file: 05_limitation.tex
\section{Limitations}\label{sec:subpopulation discussion}

One limitation of our results is that our analysis is based on empirical poisoning attacks. We measured subpopulation susceptibility using the state-of-the-art MTP and KKT attacks. While these attacks provide a lower bound the performance of poisoning attacks, it is not known how big the gap is between these attacks and the best possible poisoning attacks. Since our empirical observations are just proxies of the inherent susceptibility, future (stronger) poisoning attacks might also lead to some new insights in terms of the inherent susceptibility. 

Our analysis is limited to simple datasets and a linear SVM model, and it is not yet clear how well our observations extend to more complex models. However, as a step towards better understanding of poisoning attacks and especially in understanding how attack difficulty varies with subpopulation characteristics, experiments in such a simplified setting are valuable and revealing. Further, simple and low-capacity models are still widely used in practice due to their ease of use, low computational cost, and effectiveness \citep{tramer2020differentially,ferrari2019we}, and so our simplified analysis is still relevant in practice. Second, kernel methods or feature extraction layers in neural networks are powerful tools to handle non-linearly separable datasets by projecting them into a linearly separable high-dimensional or low-dimensional space and are widely adopted in practice. Therefore, if the important spatial relationships among the data points are still preserved after projection, then the same conclusions obtained in our simplified settings may still apply to the more complex cases by examining the spatial relationships in the transformed space. We leave these explorations as future work.

%% file: 06_conclusion.tex
\section{Summary}\label{sec:subpopulation summary}
Through extensive experiments and visualizations, we evaluate the empirical effectiveness of poisoning attacks against subpopulations in synthetic and realistic settings. We show that the difficulty of poisoning subpopulations varies widely depending on properties of the dataset and the subpopulation. Poorly separated datasets exhibit more vulnerable subpopulations in general, with subpopulation-specific properties contributing to only minor differences in vulnerability. Separable datasets, in constrast, exhibit more varied distributions of subpopulation susceptibility in which vulnerability to attack depends more heavily on the properties of the subpopulation.

We identify a strong correlation between a subpopulation's suseptibility to attack and the minimum achieveable loss difference between a model that misclassifies the subpopulation and the clean model on $\cS_c$. This correlation is generalizable, which we demonstrate on the benchmark Adult dataset. Other factors such as subpopulation size and quantities related to the margin of the clean model on the subpopulation are not by themselves strongly correlated with subpopulation susceptibility, in part due to their inability to capture information about the relative location of the subpopulation with respect to the rest of the population.

In addition to studying the general property of model loss difference, we examine the semantic properties of subpopulations from the Adult dataset and their relationship with subpopulation susceptibility. We show that significant differences in vulnerability can persist between semantically related subpopulations even when holding several numerical properties of subpopulations constant, highlight the challenges in finding generally relevant properties affecting susceptibility that are semantically meaningful.

%% file: _main.bbl
\begin{thebibliography}{10}

\bibitem{barocas_big_2016}
Solon Barocas and Andrew~D. Selbst.
\newblock Big data's disparate impact.
\newblock {\em California Law Review}, 104(3):671--732, 2016.

\bibitem{biggio2011support}
Battista Biggio, Blaine Nelson, and Pavel Laskov.
\newblock Support {V}ector {M}achines {U}nder {A}dversarial {L}abel {N}oise.
\newblock In {\em Asian Conference on Machine Learning}, 2011.

\bibitem{biggio2012poisoning}
Battista Biggio, Blaine Nelson, and Pavel Laskov.
\newblock Poisoning {A}ttacks against {S}upport {V}ector {M}achines.
\newblock In {\em International Conference on Machine Learning}, 2012.

\bibitem{buolamwini2018gender}
Joy Buolamwini and Timnit Gebru.
\newblock Gender shades: Intersectional accuracy disparities in commercial
  gender classification.
\newblock In Sorelle~A. Friedler and Christo Wilson, editors, {\em Proceedings
  of the 1st Conference on Fairness, Accountability and Transparency},
  volume~81 of {\em Proceedings of Machine Learning Research}, pages 77--91.
  PMLR, 23--24 Feb 2018.

\bibitem{carlini2023poisoning}
Nicholas Carlini, Matthew Jagielski, Christopher~A Choquette-Choo, Daniel
  Paleka, Will Pearce, Hyrum Anderson, Andreas Terzis, Kurt Thomas, and Florian
  Tram{\`e}r.
\newblock Poisoning web-scale training datasets is practical.
\newblock {\em arXiv preprint arXiv:2302.10149}, 2023.

\bibitem{chang2020adversarial}
Hongyan Chang, Ta~Duy Nguyen, Sasi~Kumar Murakonda, Ehsan Kazemi, and Reza
  Shokri.
\newblock On adversarial bias and the robustness of fair machine learning,
  2020.

\bibitem{Dua:2019}
Dheeru Dua and Casey Graff.
\newblock {UCI} {M}achine {L}earning {R}epository, 2017.

\bibitem{dwork2012fairness}
Cynthia Dwork, Moritz Hardt, Toniann Pitassi, Omer Reingold, and Richard Zemel.
\newblock Fairness through awareness.
\newblock In {\em Proceedings of the 3rd Innovations in Theoretical Computer
  Science Conference}, ITCS '12, page 214–226, New York, NY, USA, 2012.
  Association for Computing Machinery.

\bibitem{ferrari2019we}
Maurizio Ferrari~Dacrema, Paolo Cremonesi, and Dietmar Jannach.
\newblock Are we really making much progress? a worrying analysis of recent
  neural recommendation approaches.
\newblock In {\em ACM Conference on Recommender Systems}, 2019.

\bibitem{geiping2020witches}
Jonas Geiping, Liam Fowl, W~Ronny Huang, Wojciech Czaja, Gavin Taylor, Michael
  Moeller, and Tom Goldstein.
\newblock Witches' {B}rew: {I}ndustrial {S}cale {D}ata {P}oisoning via
  {G}radient {M}atching.
\newblock In {\em International Conference on Learning Representations}, 2021.

\bibitem{guyon2004nips}
Isabelle Guyon, Steve~R. Gunn, Asa Ben-Hur, and Gideon Dror.
\newblock Result analysis of the nips 2003 feature selection challenge.
\newblock {\em Advances in Neural Information Processing Systems}, 17, 2004.

\bibitem{hardt2016equality}
Moritz Hardt, Eric Price, Eric Price, and Nati Srebro.
\newblock Equality of opportunity in supervised learning.
\newblock In D.~Lee, M.~Sugiyama, U.~Luxburg, I.~Guyon, and R.~Garnett,
  editors, {\em Advances in Neural Information Processing Systems}, volume~29.
  Curran Associates, Inc., 2016.

\bibitem{huang2011adversarial}
Ling Huang, Anthony~D Joseph, Blaine Nelson, Benjamin~IP Rubinstein, and J~Doug
  Tygar.
\newblock Adversarial {M}achine {L}earning.
\newblock In {\em ACM Workshop on Security and Artificial Intelligence}, 2011.

\bibitem{huang2020metapoison}
W~Ronny Huang, Jonas Geiping, Liam Fowl, Gavin Taylor, and Tom Goldstein.
\newblock Meta{P}oison: Practical general-purpose clean-label data poisoning.
\newblock In {\em Advances in Neural Information Processing Systems}, 2020.

\bibitem{jagielski2019subpop}
Matthew Jagielski, Paul Hand, and Alina Oprea.
\newblock Subpopulation {D}ata {P}oisoning {A}ttacks.
\newblock In {\em NeurIPS 2019 Workshop on Robust AI in Financial Services},
  2019.

\bibitem{jagielski2021subpopulation}
Matthew Jagielski, Giorgio Severi, Niklas Pousette~Harger, and Alina Oprea.
\newblock Subpopulation data poisoning attacks.
\newblock In {\em Proceedings of the 2021 ACM SIGSAC Conference on Computer and
  Communications Security}, pages 3104--3122, 2021.

\bibitem{koh2017understanding}
Pang~Wei Koh and Percy Liang.
\newblock Understanding black-box predictions via influence functions.
\newblock In {\em International Conference on Machine Learning}, 2017.

\bibitem{koh2022stronger}
Pang~Wei Koh, Jacob Steinhardt, and Percy Liang.
\newblock Stronger data poisoning attacks break data sanitization defenses.
\newblock {\em Machine Learning}, 111(1):1--47, 2022.

\bibitem{mei2015security}
Shike Mei and Xiaojin Zhu.
\newblock The {S}ecurity of {L}atent {D}irichlet {A}llocation.
\newblock In {\em Artificial Intelligence and Statistics}, 2015.

\bibitem{mei2015using}
Shike Mei and Xiaojin Zhu.
\newblock Using {M}achine {T}eaching to {I}dentify {O}ptimal {T}raining-{S}et
  {A}ttacks on {M}achine {L}earners.
\newblock In {\em AAAI Conference on Artificial Intelligence}, 2015.

\bibitem{nelson2008exploiting}
Blaine Nelson, Marco Barreno, Fuching~Jack Chi, Anthony~D Joseph, Benjamin~IP
  Rubinstein, Udam Saini, Charles~A Sutton, J~Doug Tygar, and Kai Xia.
\newblock Exploiting machine learning to subvert your spam filter.
\newblock In {\em USENIX Workshop on Large Scale Exploits and Emergent
  Threats}, 2008.

\bibitem{scikit-learn}
F.~Pedregosa, G.~Varoquaux, A.~Gramfort, V.~Michel, B.~Thirion, O.~Grisel,
  M.~Blondel, P.~Prettenhofer, R.~Weiss, V.~Dubourg, J.~Vanderplas, A.~Passos,
  D.~Cournapeau, M.~Brucher, M.~Perrot, and E.~Duchesnay.
\newblock Scikit-learn: Machine learning in {P}ython.
\newblock {\em Journal of Machine Learning Research}, 12:2825--2830, 2011.

\bibitem{evan2022poisoning}
Evan Rose, Fnu Suya, and David Evans.
\newblock Poisoning attacks and subpopulation susceptibility.
\newblock In {\em 5th Workshop on Visualization for AI Explainability.}, 2022.

\bibitem{shafahi2018poison}
Ali Shafahi, W~Ronny Huang, Mahyar Najibi, Octavian Suciu, Christoph Studer,
  Tudor Dumitras, and Tom Goldstein.
\newblock Poison {F}rogs! {T}argeted clean-label poisoning attacks on {N}eural
  {N}etworks.
\newblock In {\em Advances in Neural Information Processing Systems}, 2018.

\bibitem{solans2020poisoning}
David Solans, Battista Biggio, and Carlos Castillo.
\newblock Poisoning attacks on algorithmic fairness.
\newblock {\em Joint European Conference on Machine Learning and Knowledge
  Discovery in Databases}, 2020.

\bibitem{steinhardt2017certified}
Jacob Steinhardt, Pang Wei~W Koh, and Percy~S Liang.
\newblock Certified defenses for data poisoning attacks.
\newblock {\em Advances in Neural Information Processing Systems}, 2017.

\bibitem{suya2021model}
Fnu Suya, Saeed Mahloujifar, Anshuman Suri, David Evans, and Yuan Tian.
\newblock Model-targeted poisoning attacks with provable convergence.
\newblock In {\em International Conference on Machine Learning}, 2021.

\bibitem{suya2023linear}
Fnu Suya, Xiao Zhang, Yuan Tian, and David Evans.
\newblock When can linear learners be robust to indiscriminate poisoning
  attacks?
\newblock In {\em Advances in Neural Information Processing Systems}, 2023.

\bibitem{tramer2020differentially}
Florian Tram{\`e}r and Dan Boneh.
\newblock Differentially private learning needs better features (or much more
  data).
\newblock {\em International Conference on Learning Representations}, 2021.

\bibitem{wang2022lethal}
Wenxiao Wang, Alexander Levine, and Soheil Feizi.
\newblock Lethal {D}ose {C}onjecture on data poisoning.
\newblock In {\em Advances in Neural Information Processing Systems}, 2022.

\bibitem{xiao2012adversarial}
Han Xiao, Huang Xiao, and Claudia Eckert.
\newblock Adversarial {L}abel {F}lips {A}ttack on {S}upport {V}ector
  {M}achines.
\newblock In {\em European Conference on Artificial Intelligence}, 2012.

\bibitem{zhang2017age}
Z.~Zhang, Y.~Song, and H.~Qi.
\newblock Age progression/regression by conditional adversarial autoencoder.
\newblock In {\em 2017 IEEE Conference on Computer Vision and Pattern
  Recognition (CVPR)}, pages 4352--4360. IEEE Computer Society, 2017.

\bibitem{zhu2019transferable}
Chen Zhu, W~Ronny Huang, Ali Shafahi, Hengduo Li, Gavin Taylor, Christoph
  Studer, and Tom Goldstein.
\newblock Transferable clean-label poisoning attacks on {D}eep {N}eural {N}ets.
\newblock In {\em International Conference on Machine Learning}, 2019.

\end{thebibliography}
